\documentclass[lettersize,journal]{IEEEtran}
\usepackage{amsmath,amsfonts,amssymb}
\usepackage{algorithmic}
\usepackage{algorithm}
\usepackage{array}
\usepackage{textcomp}
\usepackage{stfloats}
\usepackage{url}
\usepackage{verbatim}
\usepackage{graphicx}
\usepackage{cite}
\usepackage{subfigure}
\usepackage{booktabs}
\usepackage{bbding}
\usepackage{makecell}
\usepackage{CJKutf8}
\usepackage{etoolbox}
\makeatletter
\patchcmd{\@makecaption}
  {\scshape}
  {}
  {}
  {}
\makeatother
\usepackage{color}

\newtheorem{definition}{Definition}[section]
\newtheorem{theorem}{Theorem}

\newtheorem{lemma}{Lemma}
\newtheorem{remark}{Remark}[section]

\def\A{\mathcal{A}}
\def\B{\mathcal{B}}
\def\C{\mathcal{C}}
\def\D{\mathcal{D}}
\def\E{\mathcal{E}}
\def\F{\mathcal{F}}
\def\G{\mathcal{G}}
\def\H{\mathcal{H}}

\def\J{\mathcal{J}}
\def\K{\mathcal{K}}
\def\L{\mathcal{L}}
\def\M{\mathfrak{M}}
\def\O{\mathcal{O}}
\def\Q{\mathcal{Q}}
\def\R{\mathcal{R}}
\def\S{\mathcal{S}}
\def\T{\mathcal{T}}
\def\U{\mathcal{U}}
\def\V{\mathcal{V}}
\def\X{\mathcal{X}}

\def\Z{\mathcal{Z}}

\begin{document}
\begin{CJK}{UTF8}{gbsn}
\title{Low-Tubal-Rank Tensor Recovery via Factorized Gradient Descent}

\author{Zhiyu Liu, Zhi Han, Yandong Tang, Xi-Le Zhao, Yao Wang
\thanks{This work was supported in part by the National Natural Science Foundation of China under Grant U23A20343, 62303447, 61821005; in part by the CAS Project for Young Scientists in Basic Research under Grant YSBR-041; in part by the Liaoning Provincial ``Selecting the Best Candidates by Opening Competition Mechanism" Science and Technology Program under Grant 2023JH1/10400045;  in part by the Youth Innovation Promotion Association of the Chinese Academy of Sciences under Grant 2022196.}
\thanks{Zhiyu Liu is with the State Key
Laboratory of Robotics, Shenyang Institute of Automation, Chinese Academy
of Sciences, Shenyang 110016, P.R. China, and also with the University of Chinese Academy
of Sciences, Beijing 100049, China (email: liuzhiyu@sia.cn).}
\thanks{Zhi Han, Yandong Tang are with the State Key
Laboratory of Robotics, Shenyang Institute of Automation, Chinese Academy
of Sciences, Shenyang 110016, P.R. China (email: hanzhi@sia.cn; ytang@sia.cn).}
\thanks{Xi-Le Zhao is with the University of Electronic Science and Technology of
China, Chengdu 610051, China (e-mail: xlzhao122003@163.com).}
\thanks{Yao Wang is with the Center for Intelligent Decision-making and Machine
Learning, School of Management, Xi’an Jiaotong University, Xi’an 710049,
P.R. China. (email: yao.s.wang@gmail.com).}
\thanks{This paper has supplementary downloadable material available at http://ieeexplore.ieee.org., provided by the author. The material includes detailed proofs for Lemmas 1, 2, 3, 4, 5, 6 and auxiliary lemmas. This material is 209 kb in size.}
}



\maketitle

\begin{abstract}
This paper considers the problem of recovering a tensor with an underlying low-tubal-rank structure from a small number of corrupted linear measurements. Traditional approaches tackling such a problem require the computation of tensor Singular Value Decomposition (t-SVD), which is a computationally intensive process, rendering them impractical for dealing with large-scale tensors. Aiming to address this challenge, we propose an efficient and effective low-tubal-rank tensor recovery method based on a factorization procedure akin to the Burer-Monteiro (BM) method. Precisely, our fundamental approach involves decomposing a large tensor into two smaller factor tensors, followed by solving the problem through factorized gradient descent (FGD). This strategy eliminates the need for t-SVD computation, thereby reducing computational costs and storage requirements. We provide rigorous theoretical analysis to ensure the convergence of FGD under both noise-free and noisy situations. Additionally, it is worth noting that our method does not require the precise estimation of the tensor tubal-rank. Even in cases where the tubal-rank is slightly overestimated, our approach continues to demonstrate robust performance. A series of experiments have been carried out to demonstrate that, as compared to other popular ones, our approach exhibits superior performance in multiple scenarios, in terms of the faster computational speed and the smaller convergence error.
\end{abstract}

\begin{IEEEkeywords}
low-tubal-rank tensor recovery, t-SVD, factorized gradient descent, over parameterization.
\end{IEEEkeywords}

\section{Introduction}

Tensors are higher-order extensions of vectors and matrices, capable of representing more complex structural information in high-order data. In fact, many real-world data can be modeled using tensors, such as hyperspectral images, videos, and sensor networks. These data often exhibit low-rank properties, and hence, many practical problems can be transformed into low-rank tensor recovery problems, as exemplified by applications such as image inpainting\cite{zhang2016exact,gilman2022grassmannian,yang20223}, image and video compression\cite{wang2017compressive,baraniuk2017compressive,Wang2018SparseRF}, background subtraction from compressive measurements \cite{cao2016total,li2022tensor,Peng2022ExactDO}, computed tomography\cite{semerci2014tensor} and so on. 
The goal of low-rank tensor recovery is to recover a low-rank tensor $\X_{\star}$ from the noisy observation
\begin{equation}
    y_i=\langle \A_i, \X_{\star}\rangle + s_i , i=1,...,m, \label{equ:1.2}
\end{equation}
where $\{s_i\}^{m}_{i=1}$ is the unknown noise. This model can be concisely expressed as
\begin{equation}
    \textbf{y}=\mathfrak{M}(\X_{\star})+\textbf{s},
\notag\end{equation}
where $\mathfrak{M}(\X_{\star})=[\langle \A_1, \X_{\star}\rangle,\ \langle \A_2, \X_{\star}\rangle ,\ ...,\ \langle \A_m, \X_{\star}\rangle]$ denotes a linear compression operator and $\textbf{y}=[y_1,y_2,...,y_m]$, $\textbf{s}=[s_1,s_2,...,s_m]$. When the target tensor $\X_{\star}$ is low-rank, a straightforward method to solve this model is rank minimization
\begin{equation}
\underset{\X}{\min}\ \operatorname{rank}(\X),\ s.t. \| \textbf{y}-\mathfrak{M}(\X) \|_2\le \epsilon_s,
\label{equ:4}
\end{equation}
where $\epsilon_s\ge 0$ denotes the noise level, rank($\cdot$) is the tensor rank function. 

However, there are still numerous ambiguities in the numerical algebra of tensors. A primary concern arises from the existence of various tensor decomposition methods, each linked to its corresponding tensor rank, e.g., CONDECOMP/PARAFAC decomposition~(CP)~\cite{carroll1970analysis,harshman1970foundations}, Tucker decomposition~\cite{tucker1966some}, tensor Singular Value Decomposition (t-SVD){\cite{kilmer2011factorization}}, among others. 
The CP decomposition involves decomposing a tensor into the sum of several rank-1 out product of vectors. Hence, the CP rank corresponds to the number of rank-one decompositions employed in this factorization. While CP-rank aligns intuitively with matrix rank, its determination is rendered challenging due to its status as an NP-hard problem. Tucker decomposition initially involves unfolding the tensor into matrices along distinct modes, followed by decomposing the tensor into the product of a core tensor and the matrices corresponding to each mode. The Tucker rank corresponds to the rank of the matrices obtained after unfolding the tensor along different modes. Therefore, its rank computation is relatively straightforward, yet it cannot leverage the relationships between modes. 
Leveraging tensor-tensor product (t-product), Kilmer et al. \cite{kilmer2011factorization} introduced the t-SVD and its associated tensor rank, referred to as tubal-rank. Specifically, the t-SVD transforms a three-order tensor into the frequency domain. Subsequently, SVD is applied to each frontal slice, followed by an inverse transformation. This process yields three factor tensors, which is similar to matrix SVD. The advantage of t-SVD is that it has a theory similar to the Eckart-Young Theorem for matrices to guarantee its optimal low-tubal-rank approximation. Furthermore, the tubal-rank provides a more effective description of the low-rank property of tensors. Consequently, our emphasis in this paper lies in low-tubal-rank tensor recovery (LTRTR) under t-SVD framework.

Within the framework of t-SVD, numerous studies have already been conducted on low-rank tensor recovery. Since the rank constraint in problem (\ref{equ:4}) is NP-hard, Lu et.al\cite{lu2019tensor} proposed a t-product induced tensor nuclear norm (TNN). Hence, the optimization (\ref{equ:4}) can be loosed into a convex optimization problem \cite{lu2019tensor}
\begin{equation}
    \underset{\X\in\mathbb{R}^{n_1\times n_2 \times n_3}}{\min}\ \|\X\|_*,\ s.t.\ \| \textbf{y}-\mathfrak{M}(\X) \|_2\le \epsilon_s.
\notag\end{equation}
However, tensor nuclear norm minimization methods necessitate the computation of t-SVD decomposition, which is an extremely time-consuming process. As the dimensions of the tensor become large, the computational overhead and storage space associated with these methods become increasingly prohibitive.


\begin{table*}[]
\caption{Comparison of several LTRTR methods based on t-SVD. In column ``operator", ``measurement" means the general LTRTR problem while ``projection" denotes low-tubal-rank tensor completion problem; ``exact recovery" denotes the noiseless case while ``robust recovery" denotes the noisy case; ``over rank" denotes the over rank situation of the decomposition-based methods while ``$\backslash$" denotes that over rank situation is not applicable to the regularization-based method. The iteration complexity is calculated on a tensor in $\mathbb{R}^{n\times n \times n_3}$ with tubal-rank $r_{\star}$ while $r$ denotes the over-parameterized tubal-rank of the decomposition-based methods.}
\centering

\begin{tabular}{cccccc}
\hline
t-SVD methods & operator & recovery theory & convergence rate & iteration complexity & over rank \\ \hline
TNN \cite{lu2018exact}  & measurement               &          \Checkmark        &        \XSolidBrush          &  $\mathcal{O}(n^3n_3+n^2n_3\log n_3)$   &  $\backslash $        \\ \hline
IR-t-TNN \cite{wang2021generalized}     & measurement             &        \XSolidBrush          &       sub-linear            &   $\mathcal{O}(n^3n_3+n^2n_3\log n_3)$             &     $\backslash $      \\ \hline
RTNNM  \cite{zhang2020rip}       & measurement            &      \Checkmark           &    \XSolidBrush              &     $\mathcal{O}(n^3n_3+n^2n_3\log n_3)$                 &      $\backslash $     \\ \hline
TCTF \cite{zhou2017tensor}         &   projection                  &        \XSolidBrush         &         \XSolidBrush         &    $\mathcal{O}(r_{\star}n^2n_3+r_{\star}nn_3\log n_3)$                  &      \XSolidBrush     \\ \hline
HQ-TCSAD \cite{he2023robust} & projection    & \XSolidBrush & \XSolidBrush &   $\mathcal{O}(r_{\star}n^2n_3+r_{\star}nn_3\log n_3)$ & \XSolidBrush \\ \hline

UTF\cite{du2021unifying} & projection    & \XSolidBrush & \XSolidBrush &   $\mathcal{O}(rn^2n_3+n^2n_3\log n_3  )$ & \Checkmark \\ \hline

DNTC\&FNTC\cite{jiang2023robust} & projection    & \XSolidBrush & \XSolidBrush &   $\mathcal{O}(rn^2n_3+rnn_3\log n_3)$ & \Checkmark \\ \hline 

Ours          &   measurement                 &    \Checkmark        &  \makecell{linear $(r=r_{\star})$ \\ sub-linear $(r>r_{\star})$}                   &$\mathcal{O}(rn^2n_3+rnn_3\log n_3)$                      &    \Checkmark       \\ \hline
\end{tabular}
\label{table:1}
\end{table*}

However, a similar challenge is also present in low-rank matrix recovery. In the realm of low-rank matrix recovery, numerous endeavors have focused on decomposing large low-rank matrices into smaller factor matrices to reduce computational complexity, which is called the Burer-Monteiro (BM) method \cite{burer2003nonlinear}. Consequently, it is natural that we extend this paradigm to tensors as tensor BM factorization. To address these challenges, we propose the utilization of FGD for solving LTRTR problem. {Similar to several existing works on low-rank matrix recovery \cite{zhuo2021computational,stoger2021small,zhang2021preconditioned}, for simplicity, we consider the recovery of symmetric positive semi-definite (T-PSD) tensors. It is worth noting that with some modifications, our analytical framework can be naturally extended to more general cases.} Specifically, we decompose a symmetric positive semi-definite tensor $\X$ as $\F*\F^*$, leading to the formulation of the following optimization problem
\begin{equation}
    \underset{\F\in\mathbb{R}^{n\times r \times n_3}}{\min}\ \frac{1}{4}\|\textbf{y}-\mathfrak{M}(\F*\F^*) \|_2^2,
    \label{equ:7}
\end{equation}
where $*$ denotes the t-product (see Definition 2.4) and $\F^*$ denotes the transpose of $\F$ {(see Definition 2.5)}.
It is noteworthy that in this context, precise knowledge of the exact tubal-rank $r_{\star}$ of the original tensor is not necessary; rather, we only require that $r\ge r_{\star}$, that's over rank or over-parameterized situation. This condition proves to be more practical in real-world applications. Such an optimization problem can be efficiently solved using FGD, thereby significantly reducing computational costs. In Table \ref{table:1}, we compare our method in detail with other t-SVD based methods. In the case of exact rank, FGD demonstrates linear convergence, surpassing other methods; in the over rank situation, FGD exhibits sub-linear convergence similar to IR-t-TNN. However, it is important to highlight that the iteration complexity of FGD is significantly lower  than IR-t-TNN. In summary, our method is a general and efficient approach with theoretical underpinnings.

\subsection{Main contributions}
The main contributions of this article can be summarized as follows:
\begin{itemize}
   \item In response to the high computational complexity associated with existing methods, we propose an effective and efficient approach based on tensor BM factorization for LTRTR. Our approach involves decomposing a large tensor into the t-product of two smaller tensors, followed by the application of FGD to address the recovery problem. This method effectively reduces computational overhead. 
   \item We provide theoretical guarantees for the recovery performance of the FGD method under noiseless, noisy, exact rank, and over rank situations. In the absence of noise, our method ensures exact recovery, while in the presence of noise, it converges to a deterministic error. Furthermore, our theory ensures that, the FGD method maintains favorable performance even in the over rank scenario.
   \item The experimental outcomes affirm that our method exhibits not only the fastest convergence to the true value in noise-free scenarios but also converges to the minimum error bound in the presence of noise. Noteworthily, experimental results indicate that even when the tubal-rank is slightly overestimated, the target tensor can still be accurately recovered.
\end{itemize}

\subsection{Related work}
\subsubsection{Low rank tensor recovery with t-SVD framework}
Early works on LTRTR primarily focused on utilizing convex methods.  Lu et al. \cite{lu2019low} proposed a novel tensor nuclear norm based on reversible linear transformations and utilized nuclear norm minimization to solve the low-rank tensor completion problem. Zhang et al.\cite{zhang2020rip} pioneered the definition of the Tensor Restricted Isometry Property (T-RIP), providing proof that, the robust recovery of any third-order tensor $\X$ is attainable with the help of T-RIP. Furthermore, Zhang et al.\cite{zhang2021tensor} demonstrated that, under certain conditions on the number of measurements, linear maps satisfy the T-RIP conditions exist. In addressing the LTRTR problem under binary measurements, Hou et al.\cite{hou2021robust} introduced two t-SVD-based approaches, ensuring the successful recovery of tensors from binary measurements. While convex methods for low-rank tensor recovery offer superior experimental performance , their computational demands become prohibitive as the scale of data increases. 
In recent years, numerous efforts have been dedicated to solving the LTRTR problem using non-convex methods. These works can be primarily classified into two categories: one that transforms convex constraints into non-convex constraints, and another that decomposes large tensors into smaller tensors.  Kong et al. \cite{kong2018t} introduced a new tensor Schatten-p norm regularizer and proposed an efficient algorithm to solve the resulting nonconvex problem. However, their method does not offer rigorous recovery guarantees. Based on t-product induced tensor nuclear norm, weighted tensor nuclear norm \cite{mu2020weighted} and partial sum of tubal nuclear norm \cite{jiang2020multi} were proposed for solving the low-rank tensor completion problem. In addition, Wang et al.\cite{wang2021generalized} proposed a general non-convex method named IR-t-TNN for LTRTR. Nonetheless, their method still involves computing t-SVD, leading to a substantial computational workload. 

Another category of non-convex methods, specifically, the decomposition-based approach which decomposes large tensors into smaller tensors, has drawn a lot of attention. Zhou et al.\cite{zhou2017tensor} proposed an efficient algorithm for tensor completion based on tensor factorization avoiding the calculation of t-SVD. He et al. \cite{he2023robust} proposed a robust low-tubal-rank tensor completion method via alternating steepest descent method and tensor factorization. However, these two approaches typically necessitate knowledge of the true rank of the target tensor, a requirement that is often challenging to meet in practical situations. Some works have addressed and improved upon this issue by considering scenarios where the true rank remains unknown. Du et al. \cite{du2021unifying} introduced a Unified Tensor Factorization (UTF) method for addressing the low-tubal-rank tensor completion problem. Their approach eliminates the necessity of knowing the true tubal-rank of the original tensor. Jiang et al.\cite{jiang2023robust} based on their self-defined tensor double nuclear norm and Frobenius/nuclear hybrid norm, decomposed tensors into two factor tensors and proposed corresponding solving algorithms for tensor completion problem. However, as stated in Table \ref{table:1}, these two methods lack theoretical recovery guarantees.

\subsubsection{Low rank tensor recovery with other decomposition frameworks}
For other decomposition frameworks, a significant amount of research has been devoted to addressing the low-rank tensor recovery problem. For CP decomposition, there are various methods, including but not limited to:
sum-of-squares method for exact  and noisy \cite{barak2016noisy} tensor completion \cite{potechin2017exact}; vanilla gradient descent with nearly linear time\cite{cai2019nonconvex,cai2022uncertainty}; alternating minimization with linear convergence\cite{jain2014provable,liu2020tensor}; spectral methods \cite{montanari2018spectral,chen2021spectral,cai2021subspace}; automic norm minimization \cite{li2015overcomplete,ghadermarzy2019near}. In addition to CP decomposition, there is a substantial body of work focused on solving low-rank tensor recovery problems using Tucker decomposition, including but not limited to: nuclear norm minimization by unfolding tensor \cite{gandy2011tensor,mu2014square,huang2015provable}; spectral methods \cite{zhang2018tensor,xia2021statistically}; gradient methods on manifold optimization \cite{xia2019polynomial,cai2022generalized,wang2023implicit,luo2022tensor,tong2022scaling}. For tensor train decomposition\cite{oseledets2011tensor} and tensor ring decomposition \cite{zhao2016tensor}, the methods for low rank tensor recovery involve: parallel matrix factorization \cite{bengua2017efficient,zhang2022effective,liu2022efficient}, Riemannian preconditioned methods\cite{cai2022tensor,gao2023riemannian}, to name a few. Due to the differences in tensor decomposition frameworks, we won't delve into detailed explanations of these methods.

\section{Notations and preliminaries}
In this paper, the terms scalar, vector, matrix, and tensor are represented by the symbols $a$, $\textbf{a}$, $A$, and $\A$ respectively. {For a 3-way tensor $\A\in\mathbb{C}^{n_1 \times n_2 \times n_3}$, we denote its $(i,j,k)$-th entry as $\A(i,j,k)$ and use the Matlab notation $\A(:,:,i)$ to denote the $i$-th frontal slice. More often, the frontal slice $\A(:,:,i)$ is denoted compactly as $A^{(i)}$.} The tensor Frobenius norm is defined as $||\mathcal A||_F = \sqrt{\sum_{i,j,k}\mathcal A(i,j,k)^2}$. The spectral norm of matrices and tensors are denoted by $\|\cdot\|$. {For any tensor $\A\in\mathbb{R}^{n_1 \times n_2 \times n_3}$}, $\overline{\A}\in\mathbb{C}^{n_1\times n_2 \times n_3}$ is the Fast Fourier Transform (FFT) of $\A$ along the third dimension. In Matlab, we have $\overline{\A}=\mathtt{fft}(\A,[\ ],3)$ and $\A=\mathtt{ifft}(\overline{\A},[\ ],3)$. {The inner product of two tensors, $\mathcal A$ and $\mathcal B$, is defined as $\langle \mathcal A,\mathcal B\rangle = \sum_{i=1}^{n_3} \langle{A}^{(i)},{B}^{(i)}\rangle$. }

\begin{definition}[Block diagonal matrix \cite{kilmer2011factorization}]
For a three-order tensor $\A\in\mathbb{R}^{n_1\times n_2 \times n_3}$, we denote $\overline{A}\in\mathbb{C}^{n_1n_3\times n_2n_3}$ as a block diagonal matrix of $\overline{\A}$, i.e.,
$$
\overline{A}=\mathtt{bdiag}(\overline{\A})=\mathtt{diag}(\overline{A}^{(1)};\overline{A}^{(2)};...;\overline{A}^{(n_3)}).
$$
\end{definition}

\begin{definition}[Block circulant matrix \cite{kilmer2011factorization}]
For a three-order tensor $\A\in\mathbb{R}^{n_1\times n_2 \times n_3}$, we denote $\mathtt{bcirc}(\A)\in\mathbb{R}^{n_1n_3\times n_2n_3}$ as its block circulant matrix, i.e.,
$$
\mathtt{bcirc(\A)}=\begin{bmatrix}
A^{(1)} & A^{(n_3)} & \cdots& A^{(2)}\\
A^{(2)} & A^{(1)} & \cdots &A^{(3)}\\
\vdots & \vdots & \ddots & \vdots\\
A^{(n_3)} & A^{(n_3-1)} &\cdots & A^{(1)}\\
\end{bmatrix}.
$$
\end{definition}

\begin{definition}[The fold and unfold operations \cite{kilmer2011factorization}]
For a three-order tensor $\A\in\mathbb{R}^{n_1\times n_2 \times n_3}$, we have
\begin{equation}
    \begin{aligned}
    & \mathtt{unfold}(\A)=[A^{(1)};A^{(2)};\cdots;A^{(n_3)}]\\
&    \mathtt{fold}(\mathtt{unfold}(\A))=\A.
\end{aligned}
\notag\end{equation}
\end{definition}

\begin{definition}[T-product\cite{kilmer2011factorization}]
For $\A\in\mathbb{R}^{n_1\times n_2 \times n_3}$, $\B\in\mathbb{R}^{n_2\times q \times n_3}$, the t-product of $\A$ and $\B$ is $\mathcal{C}\in\mathbb{R}^{n_1\times q \times n_3}$, i.e., 
\begin{equation}
\C=\A*\B=\mathtt{fold}(\mathtt{bcirc}(\A)\cdot \mathtt{unfold}(\B)).
\notag
\end{equation}
\end{definition}

\begin{definition}[Tensor conjugate transpose\cite{kilmer2011factorization}]
Let $\A\in\mathbb{C}^{n_1\times n_2\times n_3}$, and its conjugate transpose is denoted as $\A^*\in\mathbb{C}^{n_2\times n_1\times n_3}$. The formation of $\A^*$ involves obtaining the conjugate transpose of each frontal slice of $\mathcal{A}$, followed by reversing the order of transposed frontal slices 2 through $n_3$. {For an example, let $\A\in\mathbb{C}^{n_1\times n_2\times 4}$ and its frontal slices be $A_1$, $A_2$, $A_3$ and $A_4$. Then 
$$
\mathcal{A}^*=\mathtt{fold}\left( \left[ \begin{array}{c}
	A_{1}^{*}\\
	A_{4}^{*}\\
	A_{3}^{*}\\
	A_{2}^{*}\\
\end{array} \right] \right),
$$
where $A^*\in\mathbb{C}^{n_2 \times n_1}$ denotes the matrix conjugate transpose of  $A\in\mathbb{C}^{n_1 \times n_2}$.}
\end{definition}

\begin{definition}[Identity tensor\cite{kilmer2011factorization}]
 The identity tensor, represented by $\mathcal{I}\in\mathbb{R}^{n\times n\times n_3}$, is defined such that its first frontal slice corresponds to the $n\times n$ identity matrix, while all subsequent frontal slices are comprised entirely of zeros. This can be expressed mathematically as:
\begin{equation}
\mathcal{I}^{(1)} = I_{n\times n},\quad \mathcal I^{(i)} = 0,i=2,3,\ldots,n_3.
\notag\end{equation}
\end{definition}

\begin{definition}[Orthogonal tensor \cite{kilmer2011factorization}]
A tensor $\mathcal{Q}\in\mathbb R^{n\times n\times n_3}$ is considered orthogonal if it satisfies the following condition:
\begin{equation}
\mathcal Q^* * \mathcal Q = \mathcal Q * \mathcal Q^*=\mathcal I.
\notag\end{equation}
\end{definition}

\begin{definition}[F-diagonal tensor \cite{kilmer2011factorization}]
A tensor is called f-diagonal if each of its frontal slices is a diagonal matrix.
\end{definition}

\begin{theorem}[t-SVD \cite{kilmer2011factorization,lu2018exact}]
Let $\A\in\mathbb{R}^{n_1\times n_2 \times n_3}$, then it can be factored as 
\begin{equation}
    \A=\U * \S * \V^*,
\notag\end{equation}
where $\U\in\mathbb{R}^{n_1 \times n_1 \times n_3}$, $\V\in\mathbb{R}^{n_2\times n_2 \times n_3}$ are orthogonal tensors, and $\S\in\mathbb{R}^{n_1\times n_2 \times n_3}$ is a f-diagonal tensor.
\end{theorem}

\begin{definition}[Tubal-rank \cite{kilmer2011factorization}]
For $\A\in\mathbb{R}^{n_1\times n_2 \times n_3}$, its tubal-rank as rank$_t(\A)$ is defined as the nonzero {diagonal} tubes of $\S$, where $\S$ is the f-diagonal tensor from the t-SVD of $\A$. That is
\begin{equation}
\operatorname{rank}_t(\A):= \#\{i:S(i,i,:)\neq 0\}.
\notag\end{equation}
\end{definition}

\begin{definition}[Tensor column space \cite{zhou2017outlier,zhang2020low}]
Consider a tensor $\X\in\mathbb{R}^{n_1\times n_2 \times n_3}$ with a tubal-rank of $r$. Let $\U*\S*\V^*$ denote the t-SVD of $\X$. The column space of $\X$ is spanned by $\U_{\X}\in\mathbb{R}^{n_1\times r \times n_3}$, where the first $r$ columns of each $\Bar{\U}_{\X}^{(i)}$ consist of the first $r$ columns of $\Bar{U}^{(i)}$, with the remaining columns being zeros. 
\end{definition}

\begin{definition}[Tensor spectral norm \cite{lu2018exact}]
For $\A\in\mathbb{R}^{n_1 \times n_2 \times n_3}$, its spectral norm is denoted as 
\begin{equation}
    \|\A\|=\|\mathtt{bcirc}(\A)\|=\|\overline{A}\|.
\notag\end{equation}
\end{definition}

{
\begin{definition}[Tensor Restricted
Isometry Property, T-RIP \cite{zhang2020rip}] A linear mapping, denoted as $\mathfrak{M}:\mathbb{R}^{n_1\times n_2 \times n_3}\to \mathbb{
R}^m$, is said to satisfy the T-RIP with parameter $(r,\ \delta_r)$ if
\begin{equation}
    (1-\delta_r)\| \X \|_F^2 \le \| \mathfrak{M}(\X) \|_2^2 \le (1+\delta_r)\| \X \|_F^2
\notag\end{equation}
holds for all tensors $\X\in \mathbb{R}^{n_1\times n_2 \times n_3}$ with a tubal-rank of at most $r$.
\end{definition}
}

\begin{definition}[Tensor condition number] The condition number of a tensor $\X\in\mathbb{R}^{n_1 \times n_2 \times n_3} $ is defined as 
    $$
    \kappa(\X)=\frac{\sigma_1(\overline{X})}{\sigma_{\min}(\overline{X})} {= \frac{\max_{k=1,2,...,n_3}\sigma_1(\overline{X}^{(k)})}{\min_{k=1,2,...,n_3}\sigma_{\min}(\overline{X}^{(k)})},}
    $$
    where {$\overline{X}$ is the block diagonal matrix of tensor $\overline{\X}$ and }$\sigma_1(\overline{X})\ \ge \sigma_2(\overline{X})\ge ...\ge \sigma_{\min}(\overline{X})>0 $ denote the singular values of $\overline{X}$. In this paper, we abbreviate $\sigma_1(\overline{X_{\star}})$ and $\sigma_{\min}(\overline{X_{\star}})$ as $\sigma_1$ and $\sigma_{\min}$, and denote $\kappa=\kappa(\X_{\star})$ for simplicity.
\end{definition}

\begin{definition}[T-positive semi-definite tensor (T-PSD) \cite{zheng2021t}]
For any symmetric $\X\in\mathbb{R}^{n \times n \times n_3}$, $\X$ is said to be a positive semi-definite  tensor, if we have
\begin{equation}
\langle \A, \X*\A \rangle \ge 0
\notag\end{equation}
for any $\A\in\mathbb{R}^{n\times 1 \times n_3}\backslash \{\mathcal{O}\}$.
\end{definition}

\section{Main results}
In this context, we initially present a detailed exposition on low-tubal-rank tensor BM factorization, followed by an analysis of its convergence rate and estimation error. Finally, we provide the algorithm's computational complexity.
\subsection{Formulation of tensor BM factorization}

As discussed in Section 1, the target of LTRTR is to recover a low-tubal-rank tensor $\X_{\star}\in\mathbb{R}^{n_1\times n_2 \times n_3}$ from the following noisy observations
\begin{equation}
    y_i=\langle \A_i, \X_{\star}\rangle + s_i , i=1,...,m, \label{equ:1.2}
    \notag
\end{equation}
where we assume $s_i\sim \mathcal{N}(0,v^2)$ is the Gaussian noise, rank$_t(\X_{\star})=r_{\star}$ and $\{\A_i\}_{i=1}^m$ are symmetric tensors with sub-Gaussian entries. Many traditional methods for LTRTR require computationally expensive t-SVD. When the tensor dimensions are large, these methods become impractical. Therefore, addressing this issue, we aim to devise an efficient and provable method for LTRTR. Drawing inspiration from the Burer-Monteiro method\cite{burer2003nonlinear}, a commonly used approach in low-rank matrix recovery, where a large matrix $X\in\mathbb{R}^{d\times d}$ is decomposed as $FF^*$, $F\in\mathbb{R}^{d\times k}$, we seek to extend this concept to the decomposition of large tensors into smaller factor tensors. 

Our first step involves investigating a tensor factorization method akin to the Burer-Monteiro method, which focuses on symmetric positive semi-definite matrix. 
For simplicity, we consider the case where the tensor is T-PSD. The more general scenarios can be extended based on our work. Subsequently, our task is to establish the proof that any T-PSD tensor $\X$ can be decomposed into $\F*\F^*$. To accomplish this, we first introduce the tensor eigenvalue decomposition based on the t-SVD framework, which is crucial for our analysis. 
\begin{theorem}(T-eigenvalue decomposition \cite{zheng2021t})
Any symmetric $\X\in \mathbb{R}^{n\times n \times n_3}$ could be decomposed into
\begin{equation}
\X=\U * \S * \U^*,
\notag\end{equation}
where $\U\in \mathbb{R}^{n\times n \times n_3 }$ is an orthogonal tensor and $\S\in\mathbb{R}^{n\times n\times n_3}$ is a f-diagonal tensor with all diagonal entries of $\overline{S}$ being the T-eigenvalues of $\X$. Note that the T-eigenvalues of $\X$ are eigenvalues of $\{X^{(i)}\}_{i=1}^{n_3}$. Especially , if $\X$ is a T-PSD tensor, then all T-eigenvalues of $\X$ are non-negative.
\end{theorem}

Based on T-eigenvalue decomposition, we derive the following lemma.
\begin{lemma}
For any symmetric T-PSD tensor $\X\in\mathbb{R}^{n\times n \times n_3}$ with tubal-rank  $r$, it can be decomposed into $\F*\F^*$, where $\F\in\mathbb{R}^{n\times r \times n_3}$.
\end{lemma}
Then we factorize a T-PSD tensor $\X$ into $\F*\F^*$, and obtain the following optimization problem:
\begin{equation}
    \underset{\F\in\mathbb{R}^{n\times r \times n_3}}{\min}\ \frac{1}{4} \|\textbf{y}-\mathfrak{M}(\F*\F^*) \|_2^2,
    \label{equ:18}
\end{equation}
where $ r_{\star}\le r \le n$, that means we do not need to accurately estimate the tubal-rank $r_{\star}$ of $\X_{\star}$. A classical approach to optimizing this problem involves obtaining an initial point close to the $\X_{\star}$, followed by the application of FGD for local convergence, which is shown in Algorithm 1. \\

{Regarding the initialization, we choose the initial point $\F_0$ as
\begin{equation}
\begin{aligned}
& \F_0 = \mathcal{P}_r\left( \M^*(\textbf{y}) \right), \M^*(\textbf{y}) = \sum_{i=1}^my_i\A_i\\ 
& \mathcal{P}_r(\M^*(\textbf{y})) = \underset{\F\in\mathbb{R}^{n\times r \times n_3}}{\operatorname{arg} \min} \|\F*\F^*-\M^*(\textbf{y})\|_F,
\end{aligned}
\label{equ:5}
\end{equation}
where the operator $\mathcal{P}_r( \cdot)$ can be calculated by T-eigenvalue decomposition, as shown in Algorithm 1. It is worth noting that similar methods have been applied in dealing with the problem of low-rank matrix recovery \cite{tu2016low,zhang2021preconditioned}, but we are the first to extend them to the t-SVD framework. Next, we shall present the theoretical guarantee to ensure that the initial point $\F_0$ obtained by (\ref{equ:5}) is close to the ground truth $\X_{\star}$.
\begin{lemma}
    Assume that $m\ge C\max\{4(2n+1)n_3r\max(v^2,1), \log(1/\epsilon)\}\delta_{4r}^{-2}$, $\delta_{4r}\le \frac{\eta\sigma_{\min} }{4\sqrt{r+r_{\star}}}$ and set the step size $\eta=\frac{1}{\rho \sigma_1}$, then we have 
    $$
    \|\F_0*\F_0^*-\X_{\star}\|_F\le \frac{1}{2\rho+4\rho\sigma_1}\sigma_{\min}
    $$
    holds with probability at least $1-\epsilon-\exp{(-\sqrt{C})}$.
\end{lemma}
}
{The above lemma indicates that when the number of samples $m$ is order optimal compared with the \textbf{degrees of freedom} of a tensor with tubal-rank $r$, the initial point $\F_0*\F_0^*$ can be close to $\X_{\star}$.
}

With this initialization assumption, we then perform FGD to solve the problem (\ref{equ:18}),
 \begin{equation}
     \F_{t+1}=\F_t-\eta\M^*(\M(\F_t*\F_t^*)-y)*\F_t.
     \label{equ:6}
\end{equation}


\begin{algorithm}[tb]
\caption{Solving~(\ref{equ:18}) by FGD}
\label{alg:trpca}
\textbf{Input:} Observation $\{y_i,\A_i\}_{i=1}^m$, step size $\eta$, estimated tubal-rank $r$ \\
{\textbf{Initialization}: Let $\U_0*\mathcal{S}_0*\U_0^*$ be the T-eigenvalue decomposition of $\M^*(\textbf{y})$, and initialize $\F_0=\sum_{i=1}^r\U_0(:,i,:)*\mathcal{S}_0(i,i,:)^{1/2}$ }
\begin{algorithmic}[1] 
\STATE \textbf{for} $t=0$ to $T-1$ \textbf{do}
\STATE \ \ \ \ \ $\F_{t+1}=\F_t-\eta\M^*(\M(\F_t*\F_t^*)-\textbf{y})*\F_t.$
\STATE \textbf{end for}
\STATE \textbf{return:} $\X_T=\F_T*\F_T^*$
\end{algorithmic}
\end{algorithm}

\subsection{Main Theorem}
We have conducted an analysis of the convergence and convergence error of FGD in four scenarios: noise-free/existence of noise, exact rank/over rank. Below, we present the direct results of our analysis. 
\begin{theorem}
Assume that $m\ge C\max\{4(2n+1)n_3r\max(v^2,1), \log(1/\epsilon)\}\delta_{4r}^{-2}$, $\delta_{4r}\le \frac{\eta\sigma_{\min} }{4\sqrt{r+r_{\star}}}$ and set step size $\eta=\frac{1}{\rho \sigma_1}$, where $\rho\ge 10$, then the ensuing claims hold with a probability of at least $1-\epsilon-\exp{(-\sqrt{C})}$:\\
\textbf{(a)} when $r=r_{\star}$ and $\textbf{s}=\textbf{0}$, we have 
$$\|\F_t*\F_t^*-\X_{\star}\|\le 0.4(1-0.8\eta\sigma_{\min})^t \sigma_{\min};$$
\textbf{(b)} when $r=r_{\star}$ and $s_i\sim \mathcal{N}(0,v^2)$, we have
$$\|\F_t*\F_t^*-\X_{\star}\|= \O\left(\kappa\sqrt{\frac{nn_3v^2}{m}}\right)$$
for any $t= \Omega\left(\frac{\log(\rho\tau/\sigma_{\min})}{\log(1-0.8\eta\sigma_{\min})}\right)$, where $\tau=\kappa\sqrt{\frac{nn_3v^2}{m}}$;\\
\textbf{(c)} when $r>r_{\star}$ and $\textbf{s}=\textbf{0}$, we have
$$
\|\F_t*\F_t^*-\X_{\star}\|\le  \frac{4c\rho\sigma_1\sigma_{\min}}{\sigma_{\min}t+10\rho c\sigma_1};
$$
\textbf{(d)} when $r>r_{\star}$ and $s_i\sim \mathcal{N}(0,v^2)$, we have
$$
\|\F_t*\F_t^*-\X_{\star}\|= \O\left(\kappa\sqrt{\frac{nn_3v^2}{m}}\right),
$$
for any $t= \Omega\left( \frac{\sqrt{m}}{\eta\kappa\sqrt{nn_3v^2}} \right)$.
\end{theorem}

\begin{remark}
In the case of exact rank ($r_{\star}=r$), regardless of the presence of noise, the convergence rate of FGD is linear. Conversely, in the over rank scenario ($r_{\star}<r$), irrespective of the existence of noise, the convergence rate of FGD is sub-linear. These findings are consistent with the experimental results presented in Section 4. 
\end{remark}

\begin{remark}
Even in the presence of noise, whether over-parameterized or not, the term $\|\F_t * \F_t^*-\X_{\star}\|$ remains constant. Regarding the recovery error $\|\F_t * \F_t^*-\X_{\star}\|_F$, we have $\|\F_t * \F_t^*-\X_{\star}\|_F \le \sqrt{r+r_{\star}}\|\F_t * \F_t^*-\X_{\star}\|$ due to the inequality $\|\X\| \le \|\X\|_F \le \sqrt{\operatorname{rank}_t(\X)}\|\X\|$. Therefore, in the case of $r=r_{\star}$, we achieve the minimal recovery error. When $r$ slightly exceeds $r_{\star}$, the impact of over rank on the recovery error is still acceptable, due to the square root factor.
\end{remark}

\begin{remark}
It is noteworthy that, among the numerous works on LTRTR, we provide, for the first time, both the convergence rate and convergence error of this problem. Convex methods such as TNN \cite{lu2018exact} and RTNNM \cite{zhang2020rip} only offer guarantees for convergence error without specifying the convergence rate. On the other hand, non-convex methods, such as the IR-t-TNN proposed by Wang et al. \cite{wang2021generalized}, provide a rough convergence rate analysis based on the Kurdyka-Lojasiewicz property. For a more detailed comparison, please refer to Table \ref{table:1}.
\end{remark}

\subsection{Population-sample analysis}
According to the T-eigenvalue decomposition, we decompose $\X_{\star}$ as
\begin{equation}
\X_{\star}=\begin{bmatrix}
    \U & \V
\end{bmatrix}*\begin{bmatrix}
    \D_{\S}^{\star} & 0 \\
    0 & 0 
\end{bmatrix} *\begin{bmatrix}
    \U & \V 
\end{bmatrix}^*,
\notag\end{equation}
where $\D_{\S}^{\star} \in \mathbb{R}^{r_{\star}\times r_{\star} \times n_3}$, $\U \in \mathbb{R}^{n\times r_{\star} \times n_3}$, $\V \in \mathbb{R}^{n\times (n-r_{\star}) \times n_3} .$ Here, $\U$ and $\V$ represent orthonormal tensors that are complementary, denoted as $\U^* * \V=0$. Consequently, for any $\F_t\in\mathbb{R}^{n\times r\times n_3}$, it can be consistently factorized as follows:
\begin{equation}
\F_t=\U*\S_t+\V*\T_t,
\notag\end{equation}
where $\S_t=\U^* * \F_t\in\mathbb{R}^{r_{\star}\times r \times n_3}$ and $\T_t=\V^**\F_t\in\mathbb{R}^{(n-r_{\star})\times r \times n_3}$.
Then we have
\begin{equation}
\begin{aligned}
\F_t*\F_t^*&=\U*\S_t * \S_t^* * \U^* + \V*\T_t*\T_t^* *\V^*\\
&\ \ \  + \U*\S_t * \T_t^* * \V^* + \V*\T_t * \S_t^* * \U^*.
\end{aligned}
\notag\end{equation}
As $t$ approaches infinity, if we have $\S_t * \S_t^*$ converges to $\D_{\S}^{\star}$, and $\S_t * \T_t^* $, $\T_t * \S_t^*$ and $ \T_t*\T_t^* $ converge to 0, then $\F_t*\F_t^*$ approximates to $\X_{\star}$.

Then we consider the gradient descent 
\begin{equation}
\begin{aligned}
    \F_{t+1}&=\F_t-\eta\M^*(\M(\F_t*\F_t^*)-\textbf{y})*\F_t \\
    &=\underbrace{\F_t-\eta(\F_t*\F_t^*-\X_{\star})*\F_t}_{\mathcal{P}_t} \\
    &-\underbrace{\left(\M^*(\M(\F_t*\F_t^*-\X_{\star})-\textbf{s})-(\F_t*\F_t^*-\X_{\star})\right)}_{\G_t}*\F_t.
\end{aligned}
\notag\end{equation}
Note that $\mathcal{P}_t$ denotes the population gradient descent while the original gradient descent denotes the sample gradient descent. Therefore, in the analysis of local convergence, we follow the typical population-sample analysis\cite{balakrishnan2017statistical,zhuo2021computational}.

\subsubsection{Population analysis}
For term $\mathcal{P}_t$, we have
\begin{equation}
\begin{aligned}
\mathcal{P}_t&=\F_t-\eta(\F_t*\F_t^*-\X_{\star})*\F_t\\
&=(\U*\S_t+\V*\T_t)-\eta\U*\D_{\S}^{\star}*\S_t\\
&\ \ \ \ -\eta(\U*\S_t+\V*\T_t)(\S_t^**\S_t+\T_t^**\T_t)\\
&=\U*\Tilde{\S_t}+\V*\Tilde{\T_t},
\end{aligned}
\notag\end{equation}
where we define
\begin{equation}
\begin{aligned}
\Tilde{\S_t}&=\S_t-\eta(\S_t*\S_t^**S_t+\S_t*\T_t^**\T_t-\D_{\S}^{\star}*\S_t)\\
\Tilde{\T_t}&=\T_t-\eta(\T_t*\T_t^**\T_t+\T_t*\S_t^**\S_t).
\end{aligned}
\notag\end{equation}
It is worth noting that $\mathcal{P}_t$ is, in fact, the gradient descent method applied to optimization problem 
\begin{equation}
    \underset{\F\in\mathbb{R}^{n\times r \times n_3}}{\min}\ \frac{1}{4} \| \F*\F^*-\X_{\star} \|_F^2.
\label{equ:10}
\end{equation}
Then we have the following Lemma:
\begin{lemma}
Suppose that we have the same settings as Theorem 3, for the case $r>r_{\star}$, we have
\begin{equation}
\begin{aligned}
&(a)     \|\D_{\S}^{\star}-\Tilde{\S_t}*\Tilde{\S_t}^*\| \\
&\ \ \ \ \ \ \ \ \ \ \ \le (1-1.7\eta\sigma_{\min})\| \D_{\S}^{\star}-\S_t*\S_t^* \| + 2.1\eta\| \S_t*\T_t^* \|^2\\
&(b) \|\Tilde{\S_t}*\Tilde{\T_t}^*\|\le  \|\S_t*\T_t^*\| (1-\eta\sigma_{\min}) \\
&(c) \|\Tilde{\T_t}*\Tilde{\T_t}^*\|\le (1-1.7\eta\|\T_t*\T_t^*\|)\|\T_t*\T_t^*\|.
\end{aligned}
\notag\end{equation}
When $r=r_{\star}$, we have 
\begin{equation}
    \|\Tilde{\T_t}*\Tilde{\T_t}^*\| \le (1-1.5\eta\sigma_{\min})\|\T_t*\T_t^*\|,
\notag\end{equation}
where the bounds for $\|\D_{\S}^{\star}-\Tilde{\S_t}*\Tilde{\S_t}^*\|$ and $\|\Tilde{\S_t}*\Tilde{\T_t}^*\|$ remain the same as (a) and (b).
\label{lemma:4}
\end{lemma}

\begin{remark}
Lemma 3 indicates that under the over-parameterized scenario, the slowest convergence is observed in $\|\Tilde{\T_t}*\Tilde{\T_t}^*\|$, which converges at a sub-linear rate, leading to sub-linear convergence in the over-parameterized setting. Conversely, in the exact rank scenario, $\|\Tilde{\T_t}*\Tilde{\T_t}^*\|$ exhibits the fastest convergence, while $\|\Tilde{\S_t}*\Tilde{\T_t}^*\|$ converges at the slowest rate in this context. However, it is noteworthy that all cases exhibit linear convergence rates in the exact rank scenario.
\end{remark}

We present an additional Lemma that establishes a connection between the population case and the sample case.
\begin{lemma}
Suppose we have the same settings as Theorem 3, then we have
\begin{equation}
\begin{aligned}
&(a)\|\D_{\S}^{\star}-\Tilde{\S_t}*\S_t^*\|\\
&\ \ \ \ \ \ \ \ \ \ \ \ \ \ \ \le(1-0.9\eta\sigma_{\min})\|\D_{\S}^{\star}-\S_t*\S_t^*\|  +\eta\|\S_t*\T_t^*\|^2\\
&(b)\|\Tilde{\S_t}*\T_t^*\|\le 1.1 \|\S_t*\T_t^*\| \\
&(c)\|\Tilde{\T_t}*S_t^*\| \le \| \T_t * \S_t^* \| \\
&(d)\| \Tilde{\T_t}*\T_t^* \| \le \|\T_t*\T_t^*\|(1-\eta\|\T_t * \T_t^*\|).
\end{aligned}
\notag\end{equation}
\end{lemma}
Building upon these two Lemmas, we proceed with the analysis under the finite sample scenario.

\subsubsection{Finite sample analysis}
For term $\G_t$, we can control it by the T-RIP and concentration inequalities. 
\begin{lemma}
Suppose that we have the same settings as Theorem 3, then we have\\
(a)$$\|\G_t\| \le \|\G_t\|_F\le \frac{1}{4}\eta\sigma_{\min}\|\F_t*\F_t^*-\X_{\star}\|$$ holds with probability at least $1-\epsilon$;\\
(b)$$\|\G_t\| \le \|\G_t\|_F\le \frac{1}{4}\eta\sigma_{\min}\|\F_t*\F_t^*-\X_{\star}\|+ \sqrt{Cnn_3\upsilon^2/m} $$ holds with probability at least  
$1-\epsilon-\exp{(-\sqrt{C})}$.
Furthermore, we have
$\sqrt{Cnn_3\upsilon^2/m}\le \frac{1}{16}\eta\sigma_{\min}$.
\label{lemma:5}
\end{lemma}

Based on the aforementioned lemmas, we can bound $\max\{\|\D_{\S}^{\star}-\S_t*\S_t^*\|,\ \|\S_t*\T_t\|,\ \|\T_t*\T_t^*\|\}$ by the following Lemma.
\begin{lemma}
Define $\E_t=\max\{\|\D_{\S}^{\star}-\S_t*\S_t^*\|, \ \|\S_t*\T_t\|, \ \|\T_t*\T_t^*\| \}$. Suppose we have the same settings as Theorem 3, then we have the following hold with high probability:\\
(a) noiseless and exact rank case,
\begin{equation}
\E_{t+1}\le (1-0.8\eta\sigma_{\min})\E_t,
\notag\end{equation}
(b) noisy and exact rank case, 
\begin{equation}
    \E_{t+1}-\frac{5\rho\kappa}{16}\sqrt{\frac{Cnn_3v^2}{m}}\le (1-0.8\eta\sigma_{\min})(\E_t-\frac{5\rho\kappa}{16}\sqrt{\frac{Cnn_3v^2}{m}}),
\notag\end{equation}
(c) noiseless and over rank case,
\begin{equation}
    \E_{t+1}\le (1-1.2\eta\E_t)\E_t,
\notag\end{equation}
(d) noisy and over rank case, 
\begin{equation}
\H_{t+1}\le (1-1.4\eta\H_t)\H_t,
\notag\end{equation} where we define $\H_t=\E_t-5\kappa\sqrt{Cnn_3\upsilon^2/m}$.\\
\end{lemma}

\begin{remark}
Lemma 6 reveals that under the exact rank scenario, irrespective of the presence of noise, $\E_t$ converges at a linear rate. On the other hand, in the over-parameterized scenario, regardless of the noise level, $\E_t$ converges at a sub-linear rate.
\end{remark}

\subsection{Main proofs}
To prove Theorem 3, it is necessary to establish all the lemmas employed in this paper. Due to space constraints, we provide the proof of only Lemma 6 here, as it is most relevant to Theorem 3. The proofs of other lemmas are deferred to the supplementary materials.
\subsubsection{Proof of Lemma 6} 
To bound $\E_t$, it is imperative to individually bound the three components $\|\D_{\S}^{\star}-\S_t*\S_t^*\|,\ \|\S_t*\T_t\|$, and $\|\T_t*\T_t^*\|$.
\leavevmode \\
\textbf{Upper bounds for $\| \S_t * \S_t^* - \D_{\S}^{\star}\|$}: 
\\
By the definitions of $\S_{t+1}$ and $\Tilde{\S}_{t+1}$, we have
\begin{equation}
\begin{aligned}
\S_{t+1}*\S_{t+1}^*&=\underbrace{\Tilde{\S_t}*\Tilde{\S_t}^*-\D_{\S}^{\star}}_{\Z_1}+\underbrace{\eta^2\U^* *\G_t* \F_t*\F_t^* * \G_t^* * \U}_{\Z_2} \\
&-\underbrace{\eta \Tilde{S_t}*\F_t^* *\G_t^* * \U}_{\Z_3} - \underbrace{\eta \U^* * \G_t * \F_t * \Tilde{S_t}^*}_{\Z_4} \\
\end{aligned}
\notag\end{equation}
For $\Z_3$ we have
\begin{equation}
\begin{aligned}
\Z_3&=\eta \Tilde{\S_t} * (\U*\S_t+\V*\T_t)^* * \G_t *\U \\
&=\eta\Tilde{\S_t} * \S_t^* * \U^* * \G_t * \U^* + \eta \Tilde{\S_t}*\T_t^* * \V^* *\G_t* \U \\
&=\underbrace{\eta\Tilde{\S_t} * \S_t^* * \U^* * \G_t * \U^* -\eta\D_{\S}^{\star}*\U^* * \G_t * \U}_{\Z_5}\\
&+\underbrace{\eta \Tilde{\S_t}*\T_t^* * \V^* *\G_t* \U}_{\Z_6}+\underbrace{\eta\D_{\S}^{\star}*\U^* * \G_t * \U}_{\Z_7}.
\end{aligned}
\notag\end{equation}
For $\Z_4$, we have 
\begin{equation}
\begin{aligned}
\Z_4&=\underbrace{\eta \U^* * \G_t * \U * \S_t * \Tilde{\S_t}^* - \eta\U^* * \G_t * \U * \D_{\S}^{\star}}_{\Z_8} \\
&+\underbrace{\eta\U^* * \G_t * \V * \T_t * \Tilde{\S_t}^* }_{\Z_9} + \underbrace{\eta\U^* * \G_t * \U * \D_{\S}^{\star}}_{\Z_{10}}.
\end{aligned}
\notag\end{equation}
Note that the spectral norm of the terms $(1)$ $\Z_5$ and $\Z_8$ are the same; $(2)$ $\Z_6$ and $\Z_9$ are the same; $(3)$ $\Z_7$ and $\Z_{10}$ are the same, which can be bounded as follows:
\begin{equation}
\begin{aligned}
\|\Z_5\|&=\|\Z_8\|\le \eta \| \D_{\S}^{\star}-\Tilde{\S_t}*\S_t^* \| \|\G_t\| \\
&\le \eta((1-0.9\eta\sigma_{\min})\|\D_{\S}^{\star}-\S_t*\S_t^*\|+\eta\|\S_t*\T_t^*\|^2)\|\G_t\|\\
&\le \eta(1-0.8\eta\sigma_{\min})\E_t\|\G_t\| \le 0.01\|\G_t\|\\
\|\Z_6\|&=\|\Z_9\|\le \eta \| \Tilde{\S_t}*\T_t^* \| \|\G_t\| \le 1.1 \eta\|\S_t*\T_t^*\| \|\G_t\|\\
&\le 0.011\|\G_t\| \\
\|\Z_7\|&=\|\Z_{10}\|\le \eta \| \D_{\S}^{\star}\| \|\G_t\|\le 0.1\|\G_t\|.
\end{aligned}
\notag\end{equation}

For term $\Z_2$, we have
\begin{equation}
\begin{aligned}
\|\Z_2\| &\le \| \eta^2\G_t* \F_t*\F_t^* * \G_t^* \| \\
& \le \eta^2 (\|\S_t*\S_t^*\|+\|\T_t*\T_t^*\|+2\|\S_t*\T_t^*\|)\|\G_t\|^2\\
&\le 1.4 \sigma_1 \eta^2 \|\G_t\|^2 \le 0.01\|\G_t\|.
\end{aligned}
\label{equ:51}
\end{equation}
Putting these results together, we obtain
\begin{equation}
\begin{aligned}
\|&\S_{t+1}*S_{t+1}-\D_{\S}^{\star}\| \\
&\le \|\Z_1 \| + \|\Z_2\| + \|\Z_7\|+\|\Z_{10}\|  \\
&\ \ \ \ + \|\Z_5\|+\|\Z_8\| + \|\Z_6\|+\|\Z_9\|\\
&\le (1-1.7\eta\sigma_{\min})\| \S_{t}*S_{t}-\D_{\S}^{\star} \|+ 0.21\eta\sigma_{\min} \E_t  + 0.25\|\G_t\| \\
&\le (1-1.2\eta\sigma_{\min})\E_t+ 0.25\sqrt{{Cnn_3\upsilon^2}/{m}} \\
\end{aligned}
\notag\end{equation}
where we using the fact that
\begin{equation}
\|\G_t\| \le  \eta\sigma_{\min} \E_t + \sqrt{Cnn_3\upsilon^2/m} .
\notag\end{equation}
and $\eta\le \frac{1}{10\sigma_1}$.\\
Consider the noiseless case, we have
\begin{equation}
 \|\S_{t+1}*\S_{t+1}^*-\D_{\S}^{\star}\| \le (1-1.2\eta\sigma_{\min})\E_t. 
\notag\end{equation}
\\
\textbf{Upper bounds for $\|\S_t * \T_t^*\|$:}\\
By the definitions of $\S_{t+1}$ and $\T_{t+1}$,  we have
\begin{equation}
\begin{aligned}
\S_{t+1} * \T_{t+1}^*&=\underbrace{\Tilde{\S_t}*\Tilde{\T_t}^*}_{\J_1}+\underbrace{\eta^2\U^* * \G_t * \F_t *\F_t^* * \G_t^* * \V}_{\J_2}\\
&+\underbrace{\eta\Tilde{\S_t}*\F_t^**\G_t^* * \V}_{\J_3}+\underbrace{\eta\U^* \G_t*\F_t*\Tilde{T_t}^*}_{\J_4}.
\end{aligned}
\notag\notag\end{equation}
For $\J_3$, we have
\begin{equation}
\begin{aligned}
\J_3&=\eta\Tilde{\S_t} * (\U*\S_t+\V*\T_t)^* * \G_t^* * \V\\
&=\underbrace{\eta\Tilde{\S_t}*\S_t^* * \U^* * \G_t^* * \V-\eta\D_{\S}^{\star}*\U^* * \G_t^* *\V}_{\J_5}\\
&+\underbrace{\eta\Tilde{\S_t}*\T_t^* * \V^* * \G_t^* * \V}_{\J_6}+ \underbrace{\eta\D_{\S}^{\star}*\U^* * \G_t^* *\V}_{\J_7}.
\end{aligned}
\notag\end{equation}
For $\J_4$, we have
\begin{equation}
\begin{aligned}
\J_4&=\eta\U^* * \G_t * (\U*\S_t+\V*\T_t) * \Tilde{\T_t}^*\\
&=\underbrace{\eta\U^* * \G_t * \S_t *\Tilde{\T_t}^*}_{\J_8}+\underbrace{\eta\U^* * \G_t * \V * \T_t *\Tilde{\T_t}^*}_{\J_9}.
\end{aligned}
\notag\end{equation}
Direct application of inequalities with operator norms leads to 
\begin{equation}
\begin{aligned}
&\|\J_5\|\le \eta \| \Tilde{\S_t}*\S_t^*-\D_{\S}^{\star} \| \|\G_t\|\le 0.01\|\G_t\|   \\
&\|\J_6\|\le \eta \| \Tilde{\S_t}* \T_t^* \| \|\G_t\| \le 0.011 \|\G_t\| \\
&\|\J_7\|\le \eta \| \D_{\S}^{\star} \| \|\G_t\| \le 0.1\|\G_t\| \\
&\|\J_8\|\le \eta \| \Tilde{\T_t}*\S_t^* \| \|\G_t\| \le 0.01\|\G_t\| \\
&\|\J_9\|\le \eta \| \Tilde{\T_t}*\T_t^* \| \|\G_t\| \le 0.01\|\G_t\|.
\end{aligned}
\notag\end{equation}
For term $\J_2$, we can bound it like term $\Z_2$ in equation (\ref{equ:51}):
\begin{equation}
\|\J_2\|=\|\Z_2\|\le  0.01\|\G_t\|.
\notag\end{equation}
Collecting all these results, we obtain
\begin{equation}
\begin{aligned}
\|&\S_{t+1}*\T_{t+1}^*\| \\
&\le \|\Tilde{\S_t}*\Tilde{\T_t}^*\| + \|\J_2\| + \eta\| \Tilde{\S_t}*\S_t^*-\D_{\S}^{\star} \| \|\G_t\| \\
&\ \ \ \ + \eta\| \Tilde{\S_t}* \T_t^* \| \|\G_t\|+  \eta \| \D_{\S}^{\star} \| \|\G_t\| +  \eta \| \Tilde{\T_t}*\S_t^* \| \|\G_t\| \\
&\ \ \ \ + \eta \| \Tilde{\T_t}*\T_t^* \| \|\G_t\| \\
&\le (1-\eta\sigma_{\min})\|\Tilde{\S_t}*\Tilde{\T_t}^*\|+ 0.2\|\G_t\| \\
&\le (1-0.8\eta\sigma_{\min})\E_t + 0.2\sqrt{{Cnn_3\upsilon^2}/{m}}
\end{aligned}
\notag\end{equation}\\
For the noiseless case, we have
\begin{equation}
\|\S_{t+1}*\T_{t+1}^*\| \le (1-0.8\eta\sigma_{\min})\E_t.
\end{equation}
\\
\textbf{Upper bounds for $\|\T_t*\T_t^*\|$:}\\
By the definition of $\T_{t+1}$, we have
\begin{equation}
\begin{aligned}
&\T_{t+1}*\T_{t+1}^* \\
&\ \ \ \ = \underbrace{\Tilde{\T_t}*\Tilde{\T_t}^*}_{\K_1}+\underbrace{\eta^2\V^* * \G_t * \F_t * \F_t^* * \G_t^* * \V}_{\K_2}\\
&\ \ \ \ +\underbrace{\eta\V^* * \G_t * \F_t * \Tilde{\T_t}^*}_{\K_3} +\underbrace{\eta\Tilde{\S_t} * \F_t^* * \G_t^* * \V}_{\K_4} \\
\end{aligned}
\notag\end{equation}
\begin{equation}
\begin{aligned}
\|\K_3\|=\|\K_4\|&=\eta \|\U^* * \G_t * (\U*\S_t+\V*\T_t)*\Tilde{\T_t}^* \| \\
&\le \eta  \| \G_t \|(\|\S_t*\Tilde{\T_t}\|+\|\|\T_t*\Tilde{\T_t}\|) \\
&\le 2.1\eta \E_t \|\G_t\|.
\end{aligned}
\notag\end{equation}
For $\K_2$, we can bound it like term $\Z_2$ in equation (\ref{equ:51}):
\begin{equation}
\|\K_2\|=\|\Z_2\|\le 0.14\eta\|\G_t\|^2.
\notag\end{equation}
Collecting all these results together, we obtain 
\begin{equation}
\begin{aligned}
\|\T_{t+1}*\T_{t+1}^*\|&\le \|\T_t*\T_t^*\|(1-1.7\eta\|\T_t * \T_t^*\|)\\
&+\|\K_2\|+2\|\K_3\| \\
&\le (1-1.7\eta\E_t)\E_t+4.2\eta\|\G_t\|^2+2.1\eta\E_t\|\G_t\| \\
&\le (1-1.4\eta\E_t)\E_t+ \eta\tau(3\E_t+4.2\tau),
\end{aligned}
\notag\end{equation}
where we take $\tau=\kappa\sqrt{Cnn_3\upsilon^2/m}$ for notation convenience.
Then we define $\H_t=\E_t-5 \tau$, and with some algebraic manipulations, we have
\begin{equation}
\|\T_{t+1}*\T_{t+1}^*\|\le (1-1.4\eta\H_t)\H_t+5\tau
\notag\end{equation}
for the noisy and over rank case. Considering the noiseless and over rank case, we obtain
\begin{equation}
    \|\T_{t+1}*\T_{t+1}^*\|\le (1-1.2\eta\E_t)\E_t.
\notag\end{equation}

For the exact parameterization case, we have 
\begin{equation}
\|\Tilde{\T_t}*\Tilde{\T_t}^*\| \le (1-1.5\eta\sigma_{\min})\|\T_t*\T_t^*\|
\notag\end{equation}
from Lemma 3. Then we have
\begin{equation}
\begin{aligned}
&\|\T_{t+1}*\T_{t+1}^*\|\\
&\ \ \ \ \le \|\T_t*\T_t^*\|(1-1.5\eta\sigma_{\min}) +\|\K_2\|+2\|\K_3\| \\
&\ \ \ \ \le \|\T_t*\T_t^*\|(1-1.5\eta\sigma_{\min}) +4.2\eta\|\G_t\|^2+2.1\eta\E_t\|\G_t\|  \\
&\ \ \ \ \le \|\T_t*\T_t^*\|(1-1.4\eta\sigma_{\min}) + 0.1 \sqrt{Cnn_3v^2/m}
\end{aligned}
\notag\end{equation}
for the noisy case. For the noiseless case, we have
\begin{equation}
\|\T_{t+1}*\T_{t+1}^*\| \le \|\T_t*\T_t^*\|(1-1.4\eta\sigma_{\min}).
\end{equation}
\\
After bounding each of the three components $\|\D_{\S}^{\star}-\S_t*\S_t^*\|,\ \|\S_t*\T_t\|$, and $\|\T_t*\T_t^*\|$ separately, we can proceed to establish bounds for $\E_t$. 
\\
\textbf{Upper bounds for $\E_t$:}\\
(a) For the noiseless and exact rank case, we have
\begin{equation}
\begin{aligned}
 \|\S_{t+1}*\S_{t+1}^*-\D_{\S}^{\star}\| &\le (1-1.2\eta\sigma_{\min})\E_t\\
 \|\S_{t+1}*\T_{t+1}^*\| &\le (1-0.8\eta\sigma_{\min})\E_t \\
    \|\T_{t+1}*\T_{t+1}^*\| & \le (1-1.4\eta\sigma_{\min})\|\T_t * \T_t^*\|.
\end{aligned}
\notag
\end{equation}
Then we obtain
\begin{equation}
\E_{t+1}\le (1-0.8\eta\sigma_{\min})\E_t.
\notag\end{equation}
(b) For the noisy and exact rank case, we have
\begin{equation}
\begin{aligned}
& \|\S_{t+1}*\S_{t+1}^*-\D_{\S}^{\star}\| \le (1-1.2\eta\sigma_{\min})\E_t+ 0.25\sqrt{{Cnn_3\upsilon^2}/{m}} \\
 &\|\S_{t+1}*\T_{t+1}^*\| \le (1-0.8\eta\sigma_{\min})\E_t + 0.2\sqrt{{Cnn_3\upsilon^2}/{m}} \\
&\|\T_{t+1}*\T_{t+1}^*\| \le  \|\T_t*\T_t^*\|(1-1.4\eta\sigma_{\min}) + 0.1 \sqrt{{Cnn_3\upsilon^2}/{m}},
\end{aligned}
\end{equation}
leading to the result
\begin{equation}
    \E_{t+1}-\frac{5\rho\kappa}{16}\sqrt{\frac{Cnn_3v^2}{m}}\le (1-0.8\eta\sigma_{\min})(\E_t-\frac{5\rho\kappa}{16}\sqrt{\frac{Cnn_3v^2}{m}}).
\notag
\end{equation}
(c) For the noiseless and over rank case, we have
\begin{equation}
\begin{aligned}
 \|\S_{t+1}*\S_{t+1}^*-\D_{\S}^{\star}\| &\le (1-1.2\eta\sigma_{\min})\E_t\\
 \|\S_{t+1}*\T_{t+1}^*\| &\le (1-0.8\eta\sigma_{\min})\E_t \\
 \|\T_{t+1}*\T_{t+1}^*\| &\le (1-1.2\eta\E_t)\E_t,
\end{aligned}
\end{equation}
leading to the result 
\begin{equation}
\E_{t+1}\le (1-1.2\eta\E_t)\E_t.
\notag
\end{equation}
(d) For the noisy and over rank case, we have
\begin{equation}
\begin{aligned}
& \|\S_{t+1}*\S_{t+1}^*-\D_{\S}^{\star}\| \le (1-1.2\eta\sigma_{\min})\E_t+ 0.25\sqrt{{Cnn_3\upsilon^2}/{m}} \\
 &\|\S_{t+1}*\T_{t+1}^*\| \le (1-0.8\eta\sigma_{\min})\E_t + 0.2\sqrt{{Cnn_3\upsilon^2}/{m}} \\
 &\|\T_{t+1}*\T_{t+1}^*\|\le (1-1.4\eta\H_t)\H_t+5\tau,
\end{aligned}
\end{equation}
where $\tau=\kappa\sqrt{Cnn_3\upsilon^2/m}$ and $\H_t=\E_t-5\tau$. Then we obtain 
\begin{equation}
\H_{t+1}\le (1-1.4\eta\H_t)\H_t.
\end{equation}
\subsubsection{Proof of Theorem 3}
Prior to establishing Theorem 3, we provide a lemma that serves to establish a connection between $\E_t$ and $\|\F_t*\F_t^*-\X_{\star}\|$ as outlined in Lemma 6.
\begin{lemma}
    Suppose that we have the same settings as
Theorem 3, define $\E_t=\max\{\|\D_{\S}^{\star}-\S_t*\S_t^*\|, \ \|\S_t*\T_t\|, \ \|\T_t*\T_t^*\| \}$,  we have 
$$
\E_t\le \|\F_t*\F_t^*-\X_{\star}\|,\ \|\F_t*\F_t^*-\X_{\star}\| \le 4\E_t.
$$
\end{lemma}
Under the assumptions of Theorem 3, Lemma 6 holds. Therefore, based on Lemma 6, we proceed to prove Theorem 3. Given that Lemma 6 provides the convergence rate and convergence error of $\E_t$, we can establish the convergence properties of $\|\F_t*\F_t^*-\X_{\star}\|$ by utilizing the relationship $\|\F_t*\F_t^*-\X_{\star}\|\le 4\E_t$. Thus, transforming the results of Lemma 6 using this relationship is sufficient to demonstrate Theorem 3.
\\
\textbf{Proof of claim (a)}\\
We have \begin{equation}
    \E_{t+1}\le (1-0.8\eta\sigma_{\min})\E_t
\notag\end{equation} from Lemma 6. 
Subsequently, it is straightforward to obtain
\begin{equation}
\E_{t} \le (1-0.8\eta\sigma_{\min})^t\E_0\le 0.1(1-0.8\eta\sigma_{\min})^t \sigma_{\min},
\notag\end{equation}
where the last inequality using the fact that $\E_0\le0.1\sigma_{\min}$. And then we obtain
\begin{equation}
\|\F_t*\F_t^*-\X_{\star}\|\le 0.4(1-0.8\eta\sigma_{\min})^t\sigma_{\min} 
\notag\end{equation}
follows from the fact that 
\begin{equation}
\begin{aligned}
& \|\F_t*\F_t^*-\X_{\star}\| \\
&\ \ \ \ \le \|\S_t*\S_t^*-\D_{\S}^{\star} \|+\| \S_t*\T_t^* \|+\|\T_t *\S_t^*\|+\|\T_t*\T_t^* \|\\
&\ \ \ \ \le 4\E_t.
\end{aligned}
\notag\end{equation}
\textbf{Proof of claim (b)}\\
We have
\begin{equation}
    \E_{t+1}-\frac{5\rho\kappa}{16}\sqrt{\frac{Cnn_3v^2}{m}}\le (1-0.8\eta\sigma_{\min})(\E_t-\frac{5\rho\kappa}{16}\sqrt{\frac{Cnn_3v^2}{m}})
\notag\end{equation}
form Lemma 6. For simplicity, we denote $\Q_t=(\E_t-\frac{5\rho\kappa}{16}\sqrt{\frac{Cnn_3v^2}{m}})$. Then we can derived that $\Q_t\le C_3\tau$ for $t\ge \frac{\log(\C_3\rho\tau/\sigma_{\min})}{\log(1-\frac{4}{5\rho\kappa})}$, leading to the result that for
$t= \Omega\left(\frac{\log(\rho\tau/\sigma_{\min})}{\log(1-\frac{4}{5\rho\kappa})}\right)$, we have$$\|\F_t*\F_t^*-\X_{\star}\|= \O\left(\kappa\sqrt{\frac{nn_3v^2}{m}}\right).$$
\\
\textbf{Proof of claim (c)}\\
We have \begin{equation}
    \E_{t+1}\le (1-1.2\eta\E_t)\E_t
\notag\end{equation}
from Lemma 6. We claim that $\E_t\le \frac{c_1}{\eta t+\frac{c_1}{\E_0}}$ holds for any $c_1\ge 1/1.2$. This holds true in the case of $t=0$. Then we assume that $\E_t\le \frac{c_1}{\eta t+\frac{c_1}{\eta\E_0}}$ holds and prove it still holds when $t=t+1$. For $t=t+1$, we have
\begin{equation}
\begin{aligned}
\E_{t+1} &\le (1-1.2\eta\E_t)\E_t \le \left(1-\frac{1.2c_1}{t+\frac{c_1}{\eta\E_0}}\right)\frac{c_1}{\eta t+\frac{c_1}{\E_0}} \\
&\le \frac{\left( t+\frac{c_1}{\eta\E_0} \right)-1.2c_1}{t+\frac{c_1}{\eta\E_0}} \frac{c_1}{\eta\left( t+\frac{c_1}{\eta \E_t} \right)}\le \frac{c_1}{\eta\left(  t+1+\frac{c_1}{\eta\E_0} \right)}.
\end{aligned}
\notag\end{equation}
Therefore we obtain $\E_t\le \frac{c_1}{\eta t+\frac{c_1}{\E_0}}$ holds for any $c_1\ge 1/1.2$ and $t\ge 0$.
Then we utilize the fact that $\|\F_t*\F_t^*-\X_{\star}\|\le 4\E_t$ to derive
\begin{equation}
\|\F_t*\F_t^*-\X_{\star}\|\le \frac{4c\rho\sigma_1\sigma_{\min}}{\sigma_{\min}t+10\rho c\sigma_1}.
\notag\end{equation}
\textbf{Proof of claim (d)}:\\
Firstly we have
\begin{equation}
\H_{t+1}\le (1-1.4\eta\H_t)\H_t
\notag\end{equation}
from Lemma 6.
We claim that $\H_t\le\frac{c_2}{\eta t + \frac{c_2}{\H_0}}$ holds for any $c_2\ge 1/1.4$, which is similar to claim (c).
Therefore, for $t=\Omega\left( \frac{\sqrt{m}}{\eta\kappa\sqrt{Cnn_3v^2}} \right)$ number of iterations, $\H_t=\E_t-5\kappa \sqrt{Cnn_3\upsilon^2/m} = \O(\kappa\sqrt{{nn_3v^2/m}})$, leading to the result
$$
\|\F_t*\F_t^*-\X_{\star}\|= \O\left(\kappa\sqrt{\frac{nn_3v^2}{m}}\right).
$$

\subsection{Complexity analysis}

Here, we analyze the time complexity of FGD. For FGD, the computational workload per iteration primarily focuses on two steps: the FFT of three order tensors and t-product. The computational complexity of performing a t-product between $\F_t\in\mathbb{R}^{n\times r \times n_3}$ and $\F_t^*\in\mathbb{R}^{r \times n \times n_3}$ is denoted as $\mathcal{O}(rn^2n_3)$. The computational complexity for calculating the FFT is denoted as $\mathcal{O}(rnn_3 \log(n_3))$. Combining the computational complexities of these two steps, we obtain the overall computational complexity per iteration for FGD, denoted as $\mathcal{O}(rn^2n_3+rnn_3\log n_3)$. As for the TNN method\cite{lu2018exact}, its computational complexity is denoted as $\mathcal{O}(n^3n_3+n^2n_3\log n_3)$. And for the complexity of other methods, please refer to Table \ref{table:1}. In comparing these methods, it is evident that our approach fully leverages the low-rank structure of the data, thereby effectively reducing the computational complexity.

\section{Numerical experiments}
We conduct various experiments to verify our theoretical results. We first conducted two sets of experiments to validate the results of Lemmas 3 and 6. Then we compare our method with a convex method TNN\cite{lu2018exact} and a non-convex method IR-t-TNN\cite{wang2021generalized} in the noiseless and noisy case. 
The experimental results indicate that our methodology not only showcases the highest computational speed in both noisy and noiseless scenarios but also attains minimal errors in the presence of noise. It is noteworthy that, even when the tubal-rank estimation for FGD is slightly higher, it still ensures excellent recovery performance. Our experiments were conducted on a laptop equipped with an Intel i9-13980HX processor (clock frequency of 2.20 GHz) and 64GB of memory, using MATLAB 2022b software.

\begin{figure*}[htbp]
\centering
\subfigure[]{
\begin{minipage}[t]{0.25\linewidth}
\centering
\includegraphics[width=4.5cm,height=4.5cm]{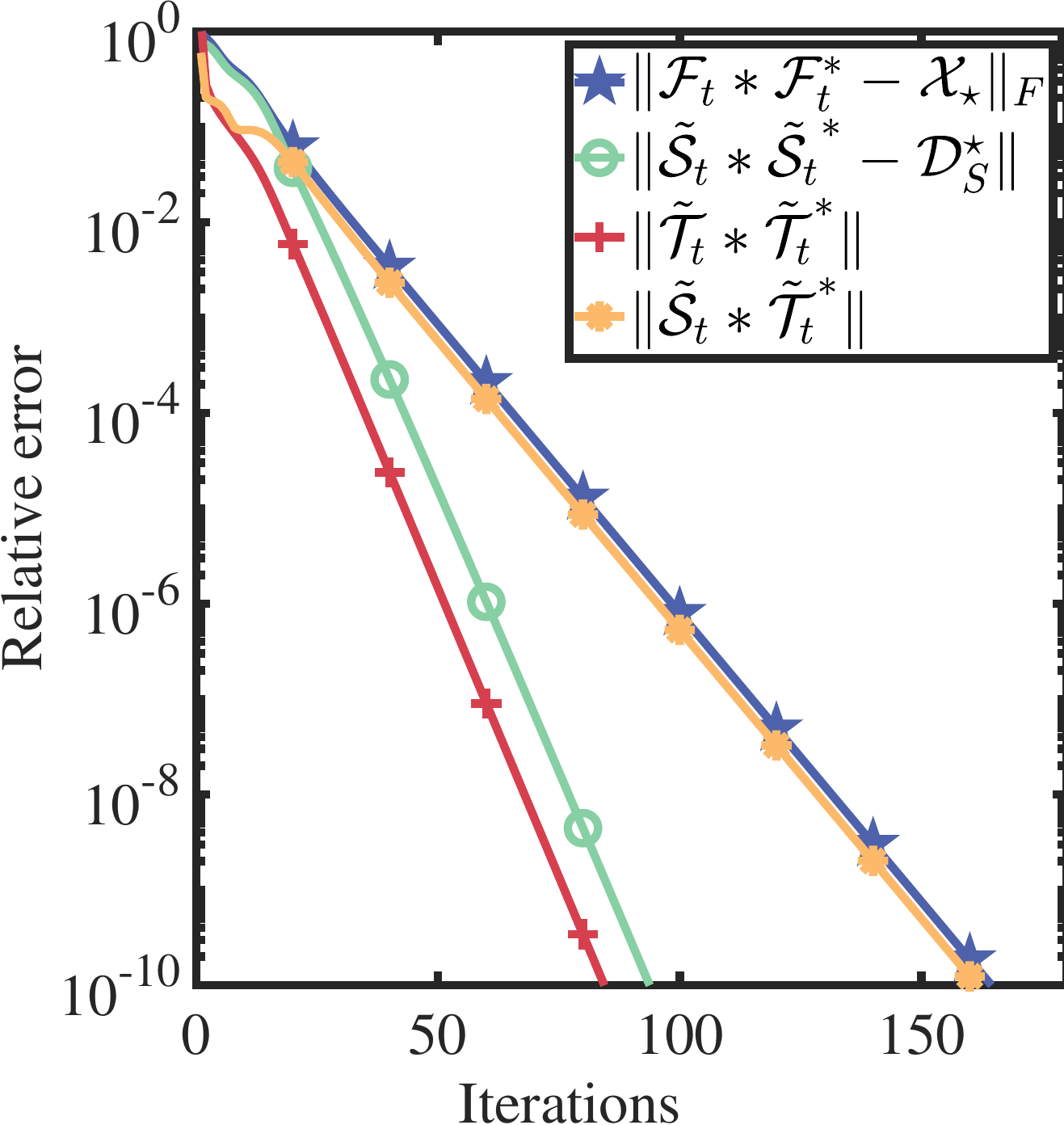}
\end{minipage}%
}%
\subfigure[]{
\begin{minipage}[t]{0.25\linewidth}
\centering
\includegraphics[width=4.5cm,height=4.5cm]{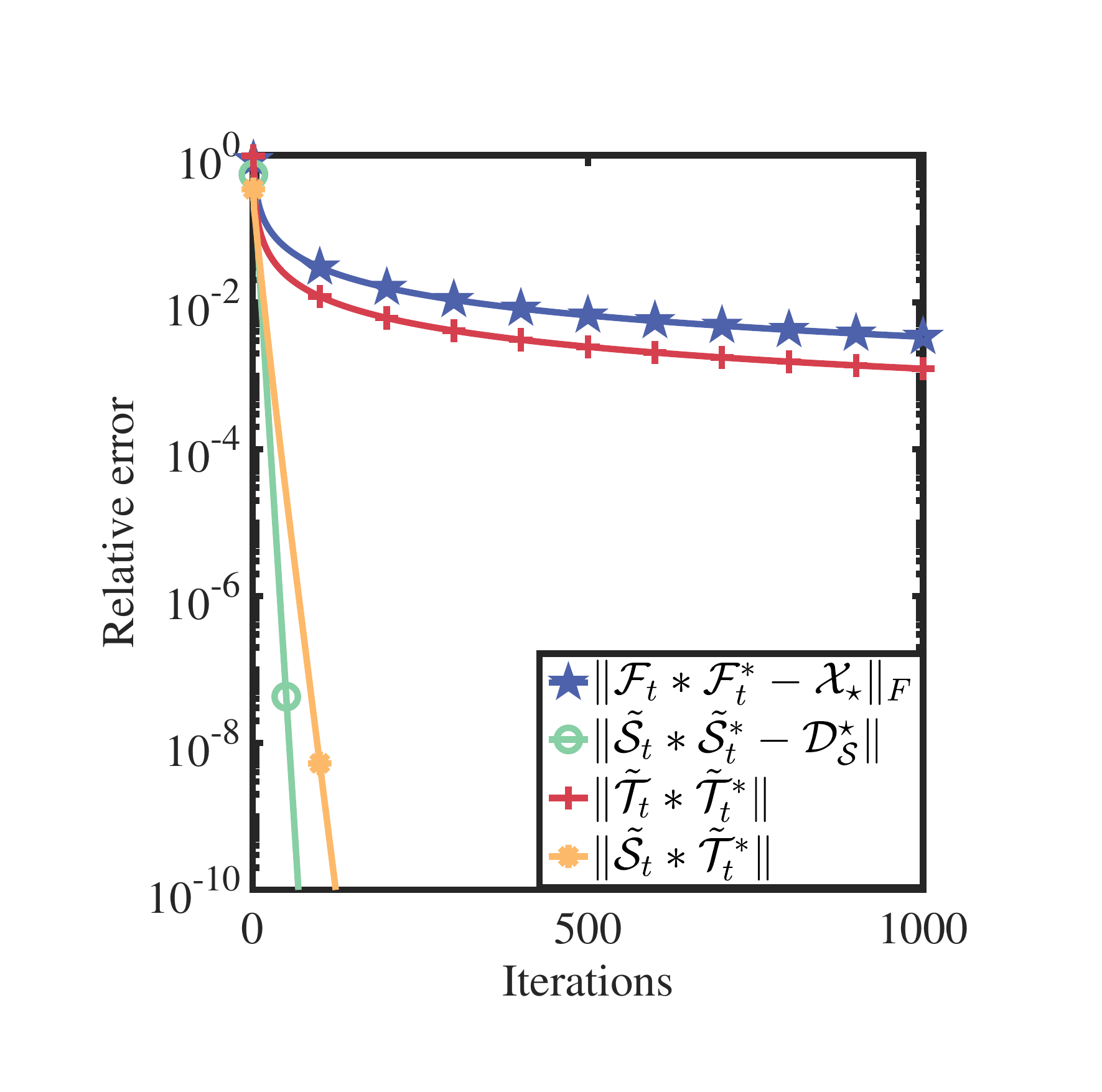}
\end{minipage}%
}%
\subfigure[]{
\begin{minipage}[t]{0.25\linewidth}
\centering
\includegraphics[width=4.5cm,height=4.5cm]{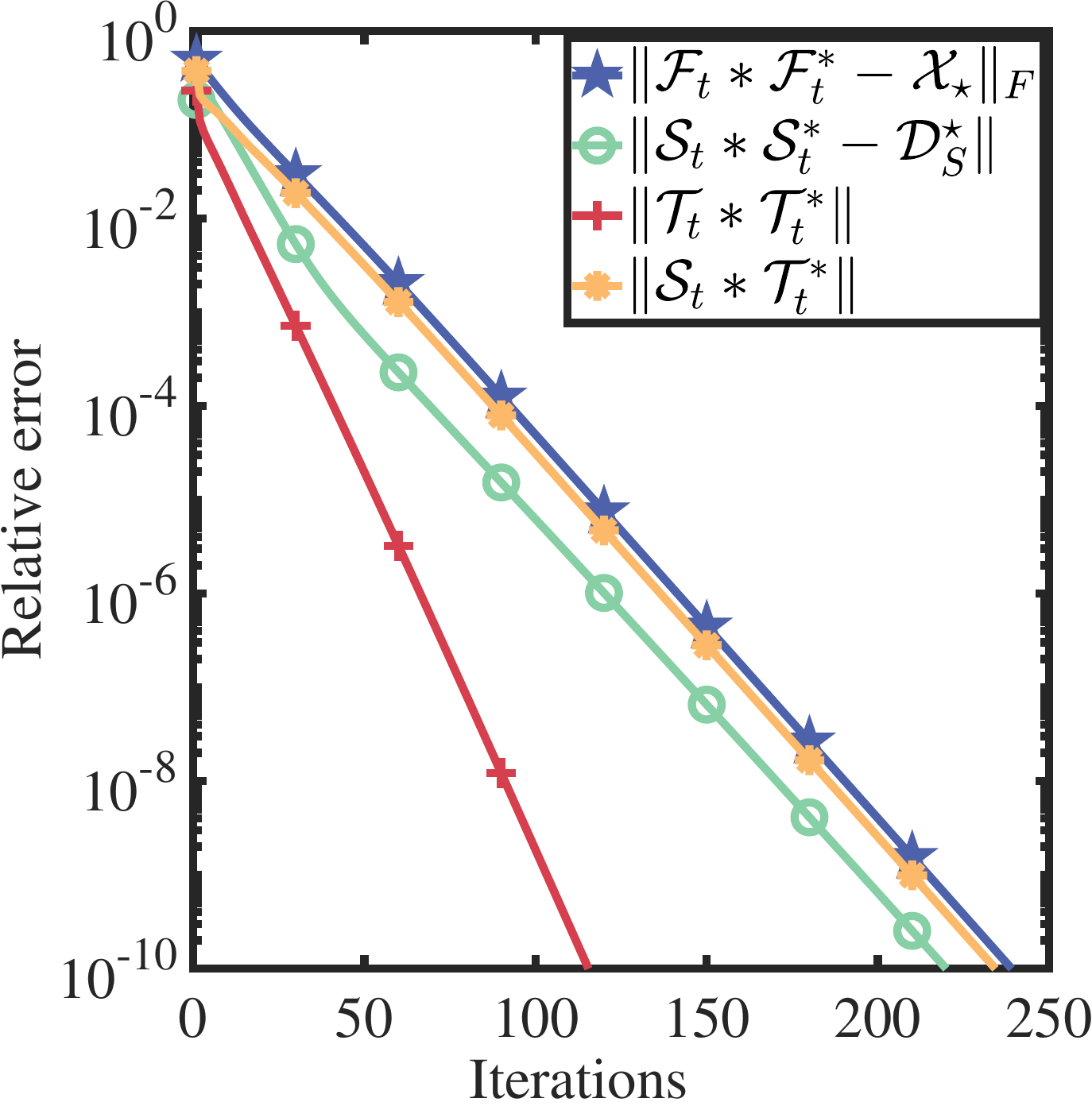}
\end{minipage}%
}%
\subfigure[]{
\begin{minipage}[t]{0.25\linewidth}
\centering
\includegraphics[width=4.5cm,height=4.5cm]{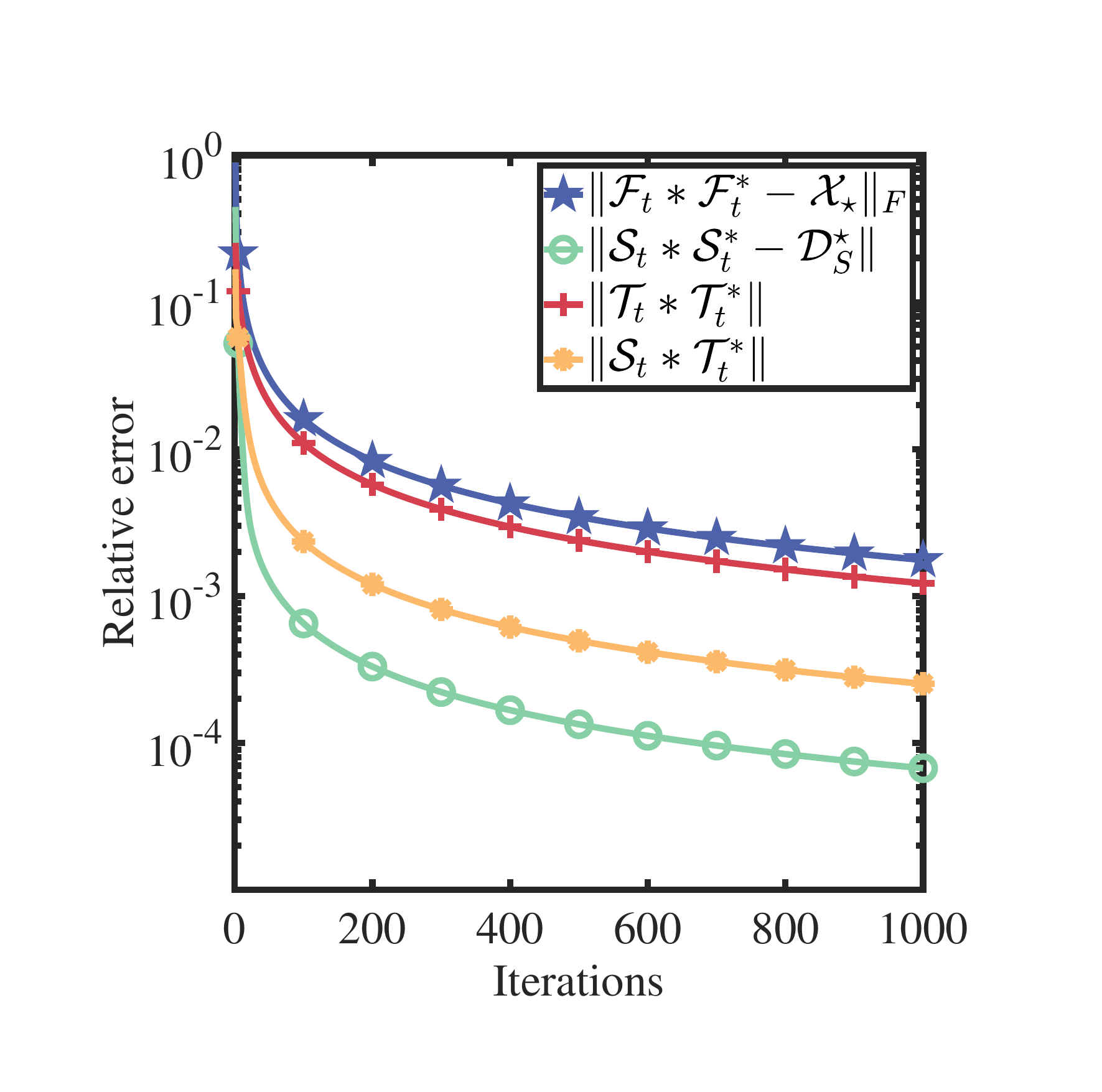}
\end{minipage}%
}%
\centering
\caption{Simulations that verify Lemmas 3 and 6. 
Subfigures (a) and (b) respectively depict the outcomes of employing FGD to solve problem (\ref{equ:10}) under the exact rank and over rank scenarios, which are the results of Lemma 3. Subfigures (c) and (d) respectively depict the outcomes of employing FGD to solve problem (\ref{equ:7}) under the exact rank and over rank scenarios, which are the results of Lemma 6. We set $n=50$, $n_3=5$, $r_{\star}=3$, $\textbf{s}=0$. For exact rank case,  we set $r=r_{\star}$; for over rank case, we set $r=5$.}

\label{fig:1}
\end{figure*}

\begin{figure}[h]
\centering
\includegraphics[scale=0.23]{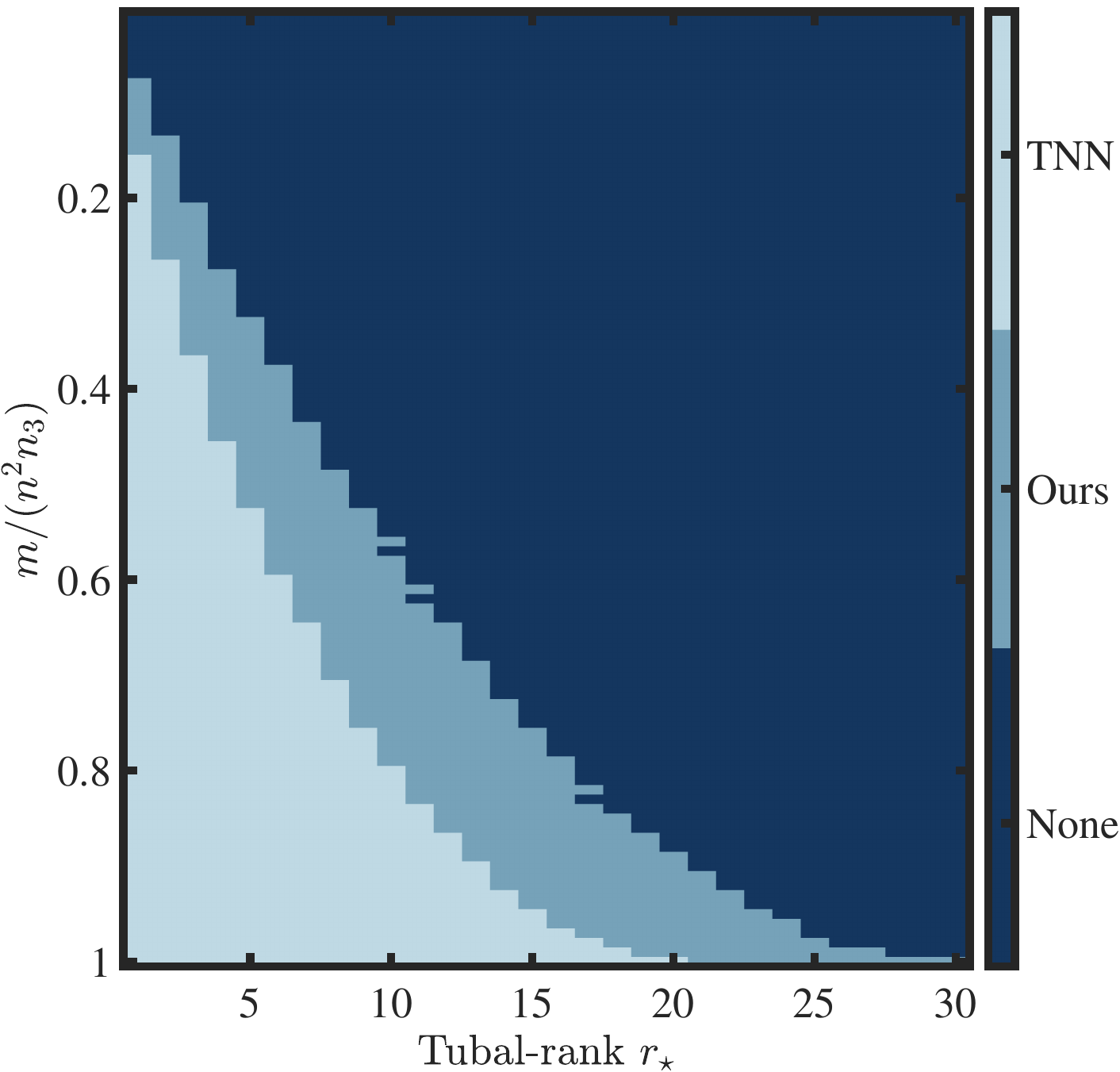}

\centering
\caption{Phase transitions for LTRTR of TNN and our FGD method. We set $n=30$, $n_3=5$. The step size of FGD is set to be 0.001. 
}
\label{fig:3}
\end{figure}

{The dataset used in this study was artificially generated based on the observation $y_i = \langle \A_i, \X_{\star}\rangle + s_i, i=1,2,...,m.$ Specifically,} 
We generate the target tensor $\X_{\star}\in\mathbb{R}^{n\times n \times n_3}$ with $\operatorname{rank}_t(\X_{\star})=r_{\star}$ by $\X_{\star}=\F_{\star}*\F_{\star}^*$, where the entries of $\F_{\star}\in\mathbb{R}^{n\times r_{\star}\times n_3}$ are i.i.d. sampled from a Gaussian distribution $\mathcal{N}(0,1)$. The entries of measurement tensor $\A_i$ are i.i.d. sampled from a Gaussian distribution $\mathcal{N}(0,\frac{1}{m})$. The entries of noise $\textbf{s}$ are i.i.d. sampled from a Gaussian distribution $\mathcal{N}(0,v^2)$. {For the FGD method, the initial value $\F_0$ is obtained according to Algorithm 1; for the other two methods, their initial values are obtained according to the description in the original paper}. The parameters of TNN, IR-t-TNN, and FGD are fine-turned for the best result.

\subsection{Verify the convergence rate}
 For the results of Lemma 3, we conducted two sets of experiments to validate the convergence rates under the scenarios of exact rank and over rank. As depicted in Figure 1 (a) and (b), it is evident that under the exact rank scenario, convergence occurs at a linear rate. Conversely, in the over rank scenario, $\|\Tilde{\T_t}*\Tilde{\T_t}^*\|$ and $\|\F_t*\F_t^*-\X_{\star}\|_F$ exhibit sub-linear convergence, with $\|\Tilde{\T_t}*\Tilde{\T_t}^*\|$ determining the rate of convergence for $\|\F_t*\F_t^*-\X_{\star}\|_F$. Meanwhile, $\|\Tilde{\S_t}*\Tilde{\T_t}^*\|$ and $\|\D_{\S}^{\star}-\Tilde{\S_t}*\Tilde{\S_t}^*\|$ converge at a linear rate in the over rank scenario.

Regarding Lemma 6, we conducted two sets of noise-free experiments to verify the convergence rates under the scenarios of exact rank and over rank. As illustrated in Figure 1 (c) and (d), it is evident that under the exact rank scenario, convergence is linear, while under the over rank scenario, convergence is sub-linear. In comparison to Lemma 3, in the over-parameterized scenario, $\|\T_t*\T_t^*\|$ still determines the convergence of $\|\F_t*\F_t^*-\X_{\star}\|_F$. However, unlike Lemma 3, under the over-parameterized scenario, both $\|\D_{\S}^{\star}-\S_t*\S_t^*\|$ and $\|\S_t*\T_t\|$ exhibit sub-linear convergence。

\subsection{Noiseless case}

For the noiseless scenario, we conducted two sets of experiments comparing the recovery performance and computational efficiency of FGD, TNN, and IR-t-TNN under exact rank and over rank situations. From Figure 1, it can be observed that, under exact rank ($r=r_{\star}$) situation, all three methods can ensure exact recovery. It is noteworthy that our approach achieves a linear convergence rate with minimal computational cost, consistent with our theoretical expectations. In the case of over rank ($r>r_{\star}$), our method exhibits sub-linear convergence but still guarantees convergence to a small relative error within a relatively short time. Furthermore, when the error reaches 0.01, the number of iterations required by our method and TNN is comparable, but the computation time for our method is significantly smaller than that for TNN, indicating the lower computational complexity of each step in our approach.

Additionally, we investigated the phase transition phenomena of FGD and TNN under different tubal-rank $r_{\star}$ and the number of measurements $m$. The IR-t-TNN algorithm, requiring a larger number of measurements, exhibits relatively poor performance in this experiment; hence, it is not included in the comparison. We set $n=30$, $n_3=5$. We vary $m$ between $0.01n^2n_3$ and $n^2n_3$, and vary $r_{\star}$ between 1 and 30. For each pair of $(m, r)$, we conducted 10 repeated experiments. In each experiment, we set the maximum number of iterations to 1000, and when the relative error $\|\hat{\X}-\X_{\star}\|_F/\|\X_{\star}\|_F\le 10^{-2}$, we considered $\X_{\star}$ to be successfully recovered. When the number of successful recoveries in 10 repeated experiments is greater than or equal to 5, we categorize that experimental set as a successful recovery, as illustrated in the graph. The deep blue region represents unsuccessful recoveries, while the other two colors represent the recovery outcomes of our method and TNN. It is evident that our method demonstrates significantly superior recovery performance compared to the TNN algorithm.

\subsection{Noisy case}
For the noisy scenario, we initially investigated the convergence rate and convergence error of FGD under both exact rank and over rank situations. From Figure 4, it is evident that FGD converges to the minimum relative error at the fastest rate, regardless of whether it is under exact rank or over rank conditions. In the case of exact rank, the convergence is linear, while in the over rank situation, it is sub-linear. In contrast, the final convergence errors for the other two methods are larger than that of FGD. Specifically, TNN converges initially to an error comparable to FGD and then diverges to a larger error. Additionally, it is observed that the computation time for IR-t-TNN is significantly smaller than that for TNN, demonstrating the efficiency of the non-convex method.

\begin{figure*}[htbp]
\centering
\subfigure[]{
\begin{minipage}[t]{0.25\linewidth}
\centering
\includegraphics[width=4.5cm,height=4.5cm]{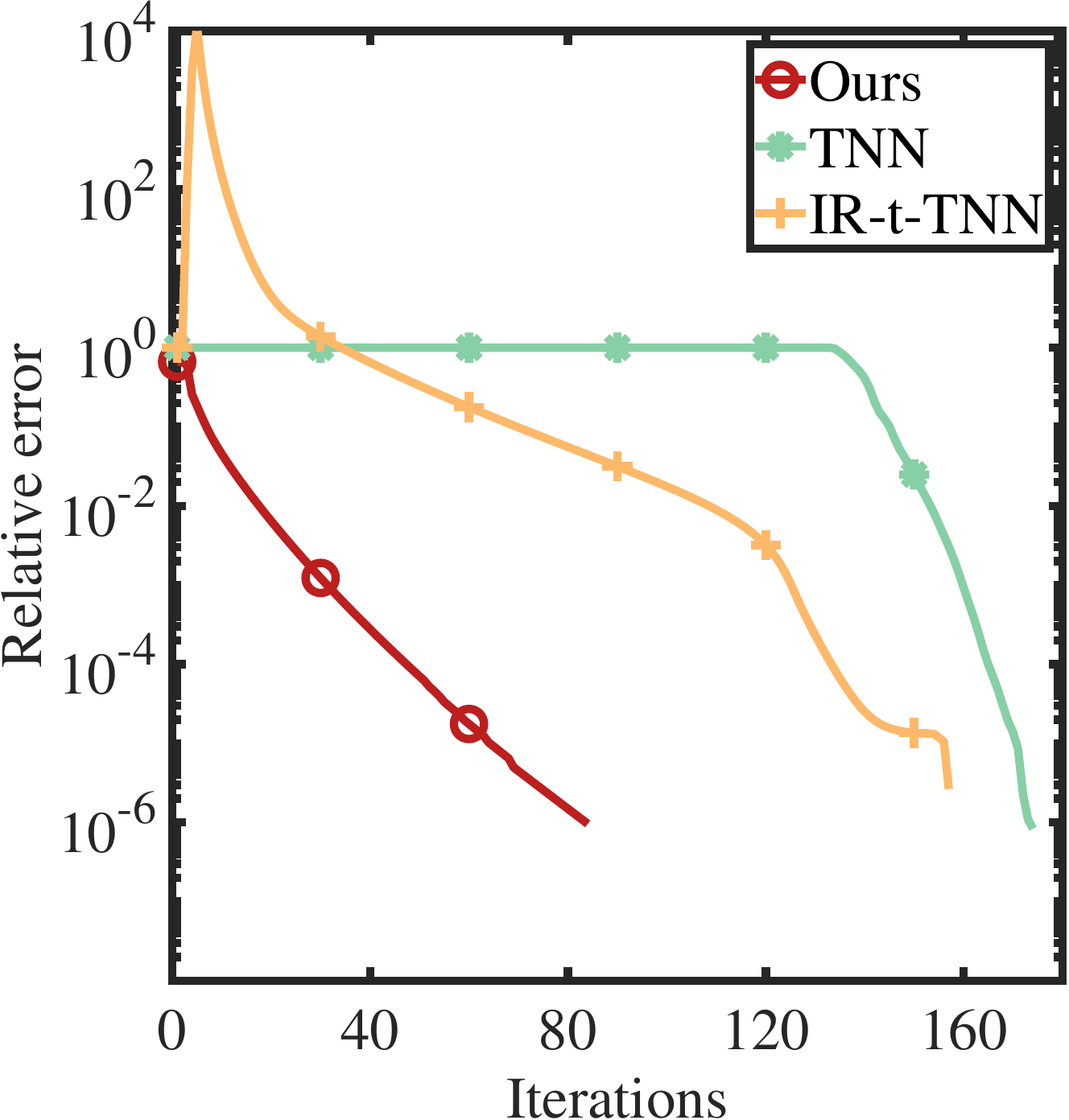}
\end{minipage}%
}%
\subfigure[]{
\begin{minipage}[t]{0.25\linewidth}
\centering
\includegraphics[width=4.5cm,height=4.5cm]{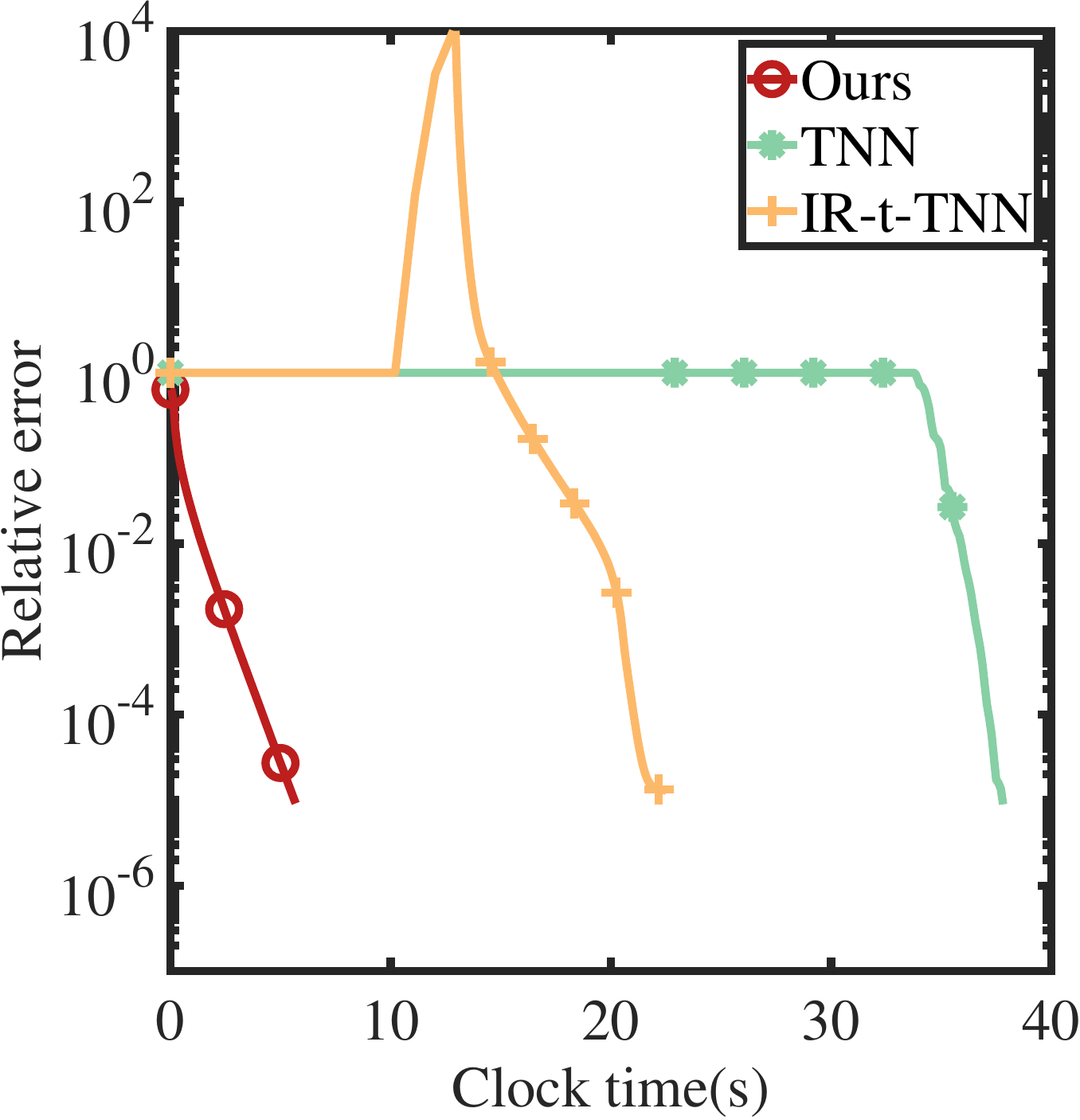}
\end{minipage}%
}%
\subfigure[]{
\begin{minipage}[t]{0.25\linewidth}
\centering
\includegraphics[width=4.5cm,height=4.5cm]{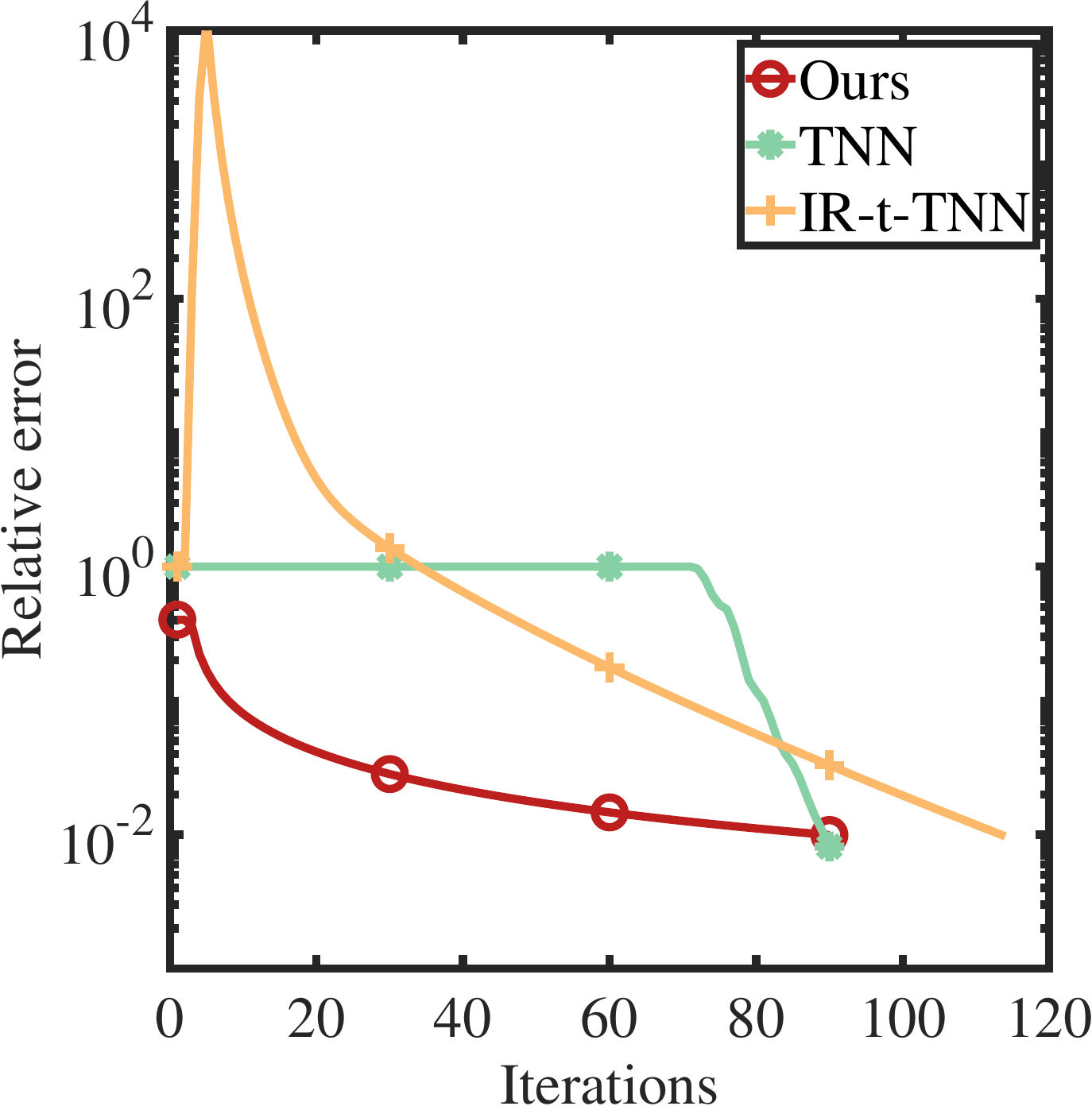}
\end{minipage}%
}%
\subfigure[]{
\begin{minipage}[t]{0.25\linewidth}
\centering
\includegraphics[width=4.5cm,height=4.5cm]{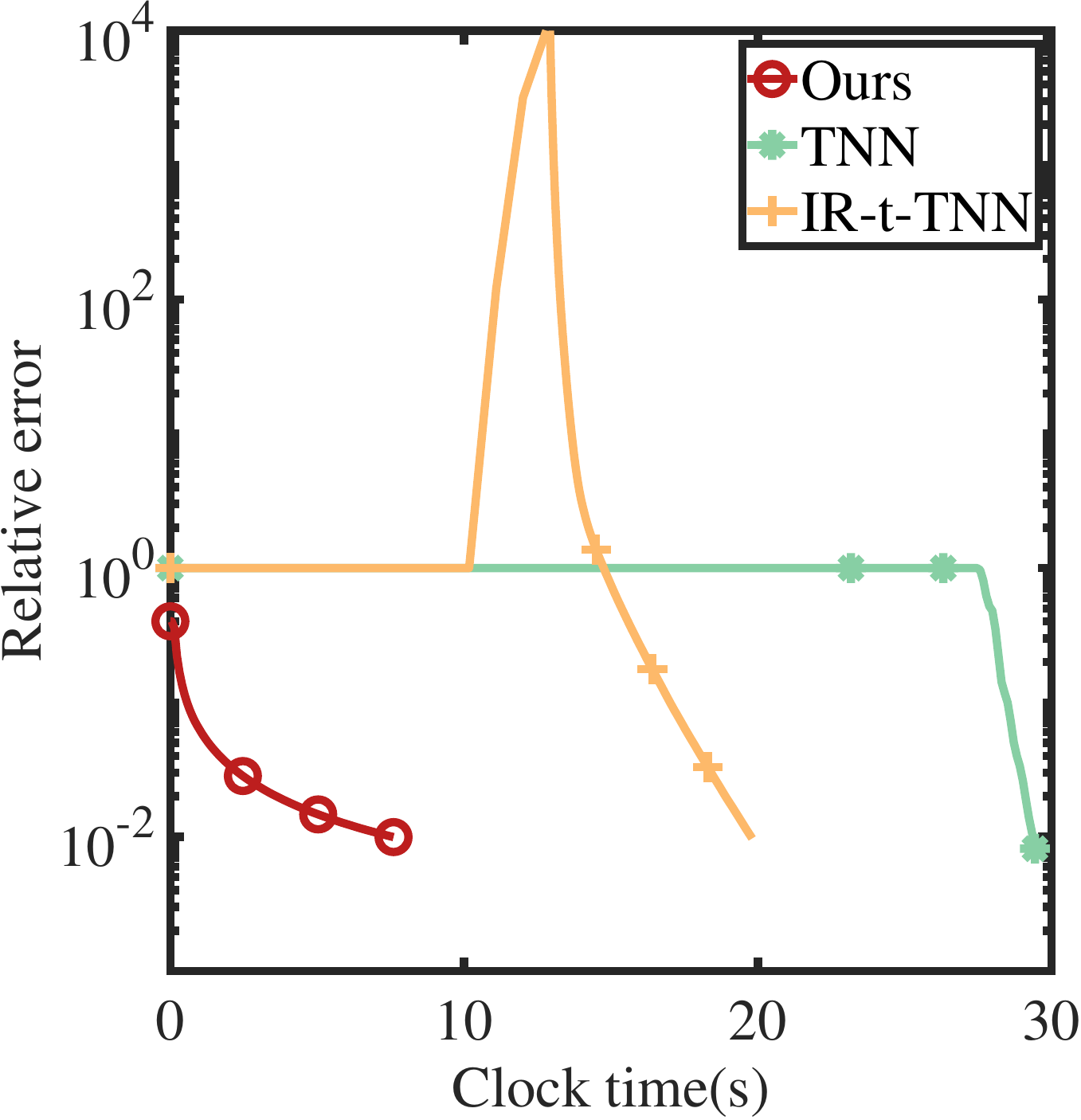}
\end{minipage}%
}%
\centering
\caption{The relative error and clock time of three methods in the noiseless case. Subfigures (a) and (b) present a comparison of the convergence rate and computation time for three algorithms under the exact rank scenario. Subfigures (c) and (d) present a comparison of the convergence rate and computation time for three algorithms under the over rank scenario. We set $n=50,\ n_3=5,\ r_{\star}=3,\ m=10(2n-r_{\star})n_3,\ \eta=0.001$. For exact rank case, we set $r=r_{\star}$; for over rank case, we set $r=r_{\star}+2$. We stop when $\|\X_t-\X_{\star}\|_F/\|\X_{\star}\|_F\le 10^{-5}$ for exact rank case and $\|\X_t-\X_{\star}\|_F/\|\X_{\star}\|_F\le 10^{-2}$ for the over rank case.
}
\label{fig:2}
\end{figure*}

\begin{figure*}[htbp]
\centering
\subfigure[]{
\begin{minipage}[t]{0.25\linewidth}
\centering
\includegraphics[width=4.5cm,height=4.5cm]{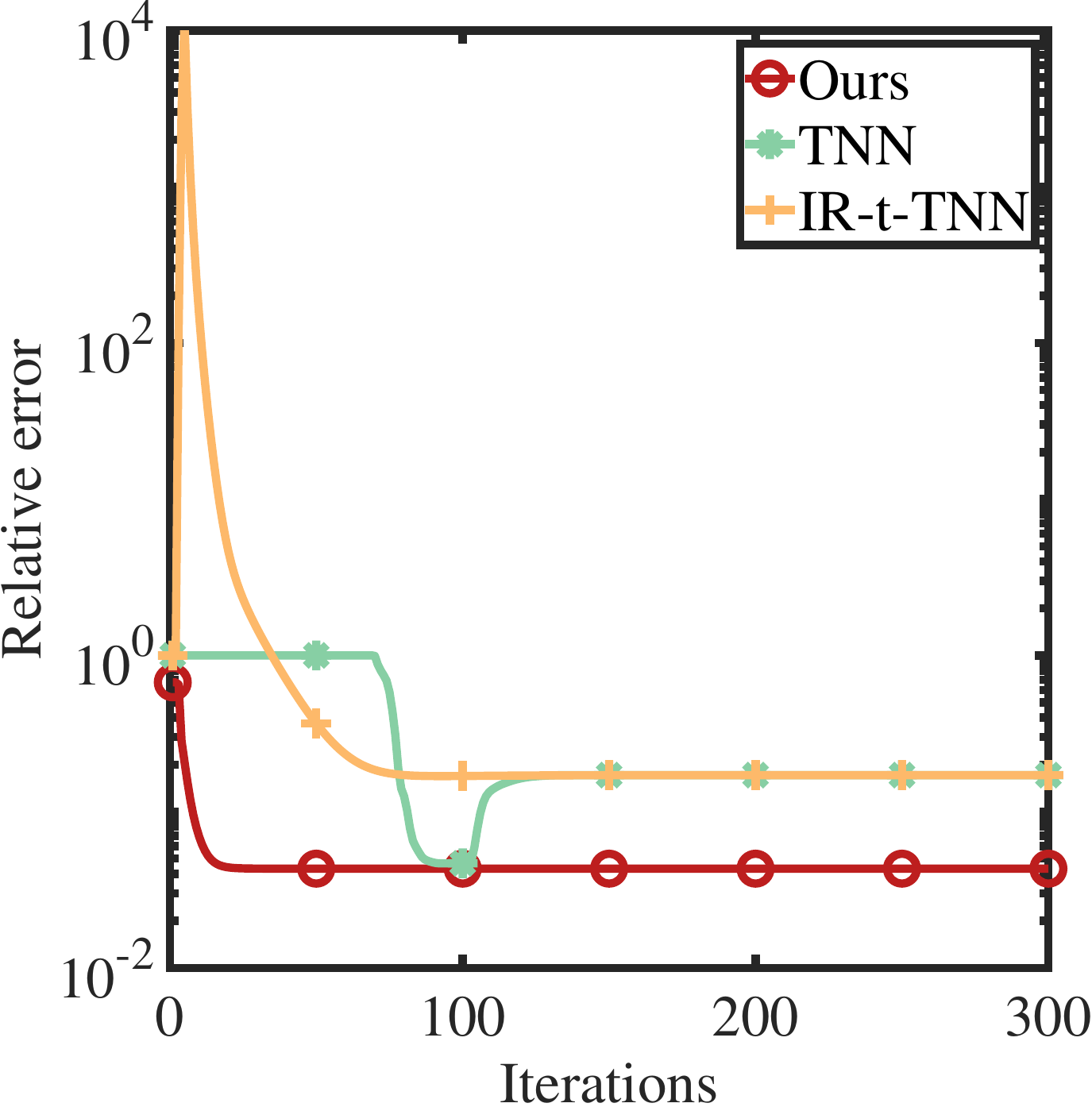}
\end{minipage}%
}%
\subfigure[]{
\begin{minipage}[t]{0.25\linewidth}
\centering
\includegraphics[width=4.5cm,height=4.5cm]{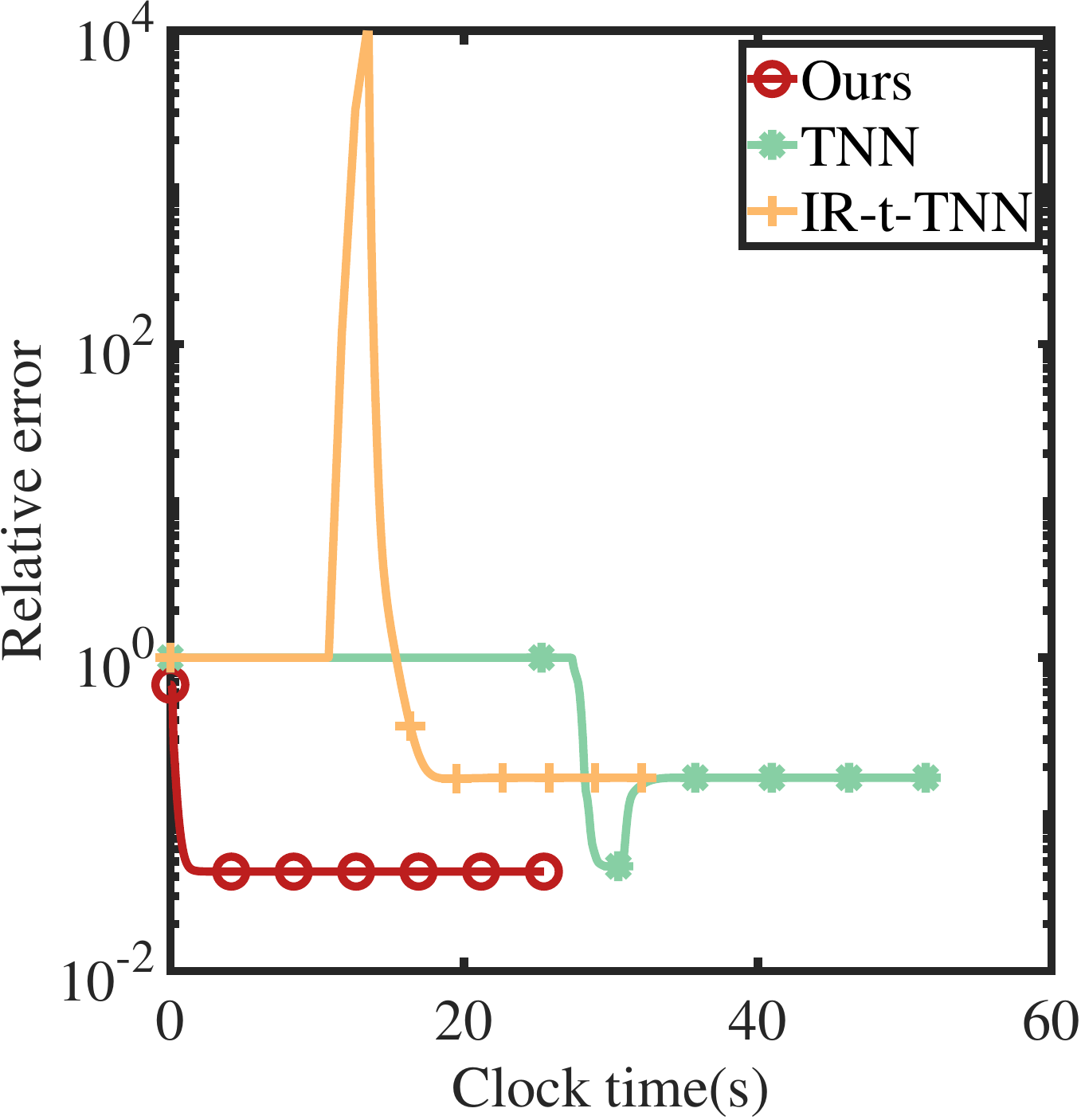}
\end{minipage}%
}%
\subfigure[]{
\begin{minipage}[t]{0.25\linewidth}
\centering
\includegraphics[width=4.5cm,height=4.5cm]{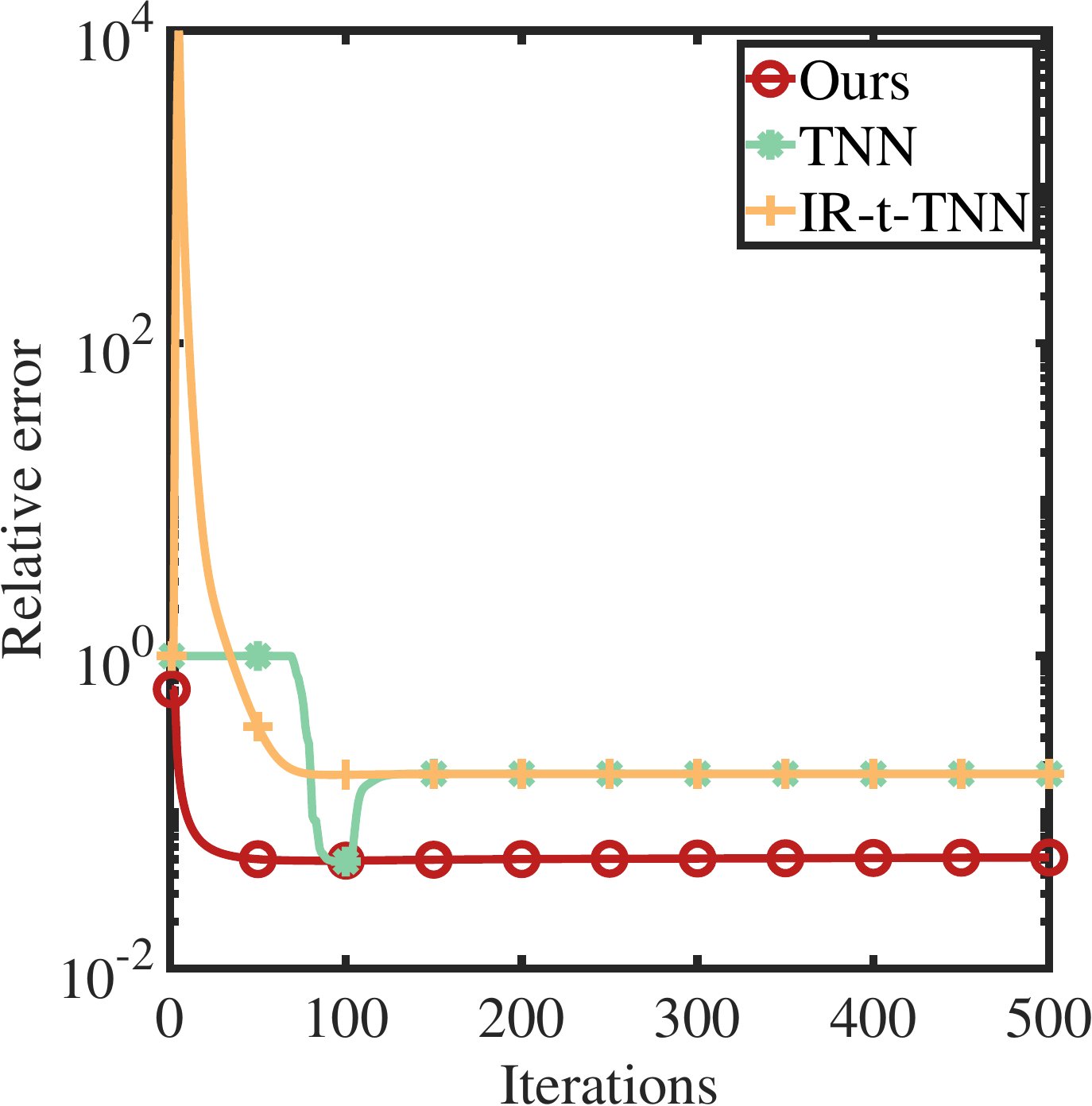}
\end{minipage}%
}%
\subfigure[]{
\begin{minipage}[t]{0.25\linewidth}
\centering
\includegraphics[width=4.5cm,height=4.5cm]{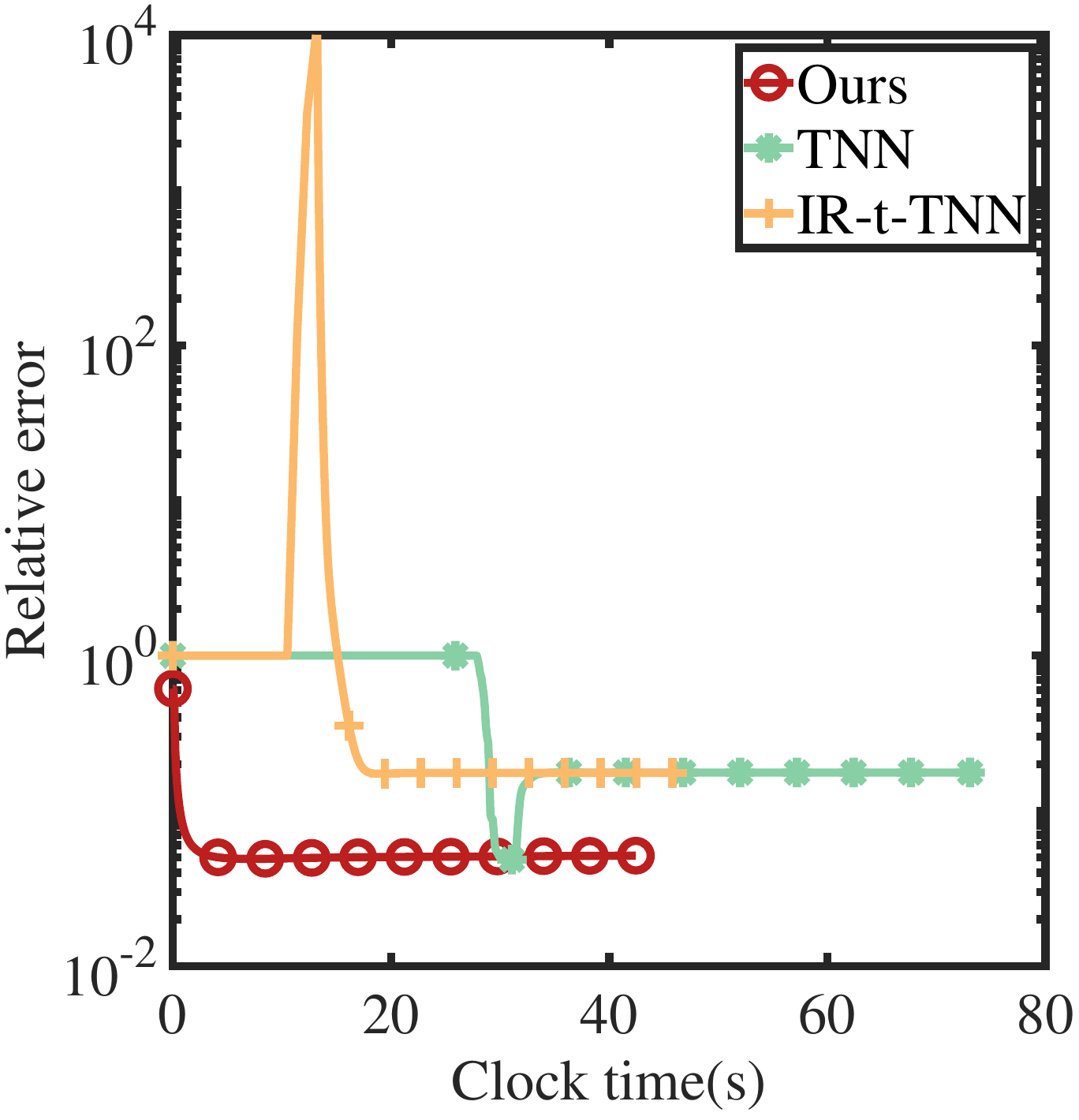}
\end{minipage}%
}%
\centering
\caption{The relative error and clock time of three methods in the noisy case. Subfigures (a) and (b) present a comparison of the convergence rate and computation time for three algorithms under the exact rank scenario. Subfigures (c) and (d) present a comparison of the convergence rate and computation time for three algorithms under the over rank scenario. We set $n=50,\ n_3=5,\ r_{\star}=3,\ m=10(2n-r_{\star})n_3,\ s_i\sim\mathcal{N}(0,0.5^2),\ \eta=0.001$. For exact rank case, we set $r=r_{\star}$; for over rank case, we set $r=r_{\star}+2$. 
}
\label{fig:3}
\end{figure*}

Then we conducted a series of experiments considering various values of $n$, tubal-rank, and noise, with the experimental results presented in Tables \ref{table:2}, \ref{table:3} and \ref{table:4}. The stopping criteria for the TNN and IR-t-TNN algorithms were consistent with the original papers. The measurements number $m$ is set to be $10r_{\star}n_3(2n-r_{\star})$. For our algorithm, the termination point was set as $\frac{\|\X_{t+1}-\X_t\|_F}{\|\X_t\|_F}\le 5 \times 10^{-4}$, and $\operatorname{rank}_t(\X)=r_{\star}+2$. All experiments in the tables are repeated 10 times.

Through these three sets of experiments, the following conclusions can be drawn:
\begin{itemize}
    \item In all scenarios of the three experiments, our FGD method achieved the minimum relative recovery error with the shortest computation time. For TNN and IR-t-TNN, their final recovery errors are very close. Additionally, the non-convex IR-t-TNN method has a significantly smaller computation time than the convex TNN method but is still much larger than our FGD method.
    
    \item As the variance of the noise increases, the recovery errors for all three methods also increase. This aligns with the theoretical expectations of recovery error growth with higher noise levels.
    
    \item As \(r_{\star}\) increases, the recovery errors for all three methods decrease. This is because the measurement number \(m\) increases with the growth of \(r_{\star}\), leading to a reduction in recovery errors, consistent with our theoretical expectations. Furthermore, with the increase in \(r_{\star}\), the computation time for all three methods significantly increases.
    
    \item As \(n\) increases, the recovery errors for all three methods noticeably decrease. This is because the increase in \(n\) results in an increase in the measurement number \(m\), consistent with theoretical expectations.
\end{itemize}

\section{Conclusion}
This article proposes a non-convex LTRTR method based on a tensor type of  Burer-Monteiro factorization, which significantly reduces computational and storage costs as compared with existing popular methods. We provide rigorous convergence guarantees and demonstrate that our method performs very well in both exact rank and over rank scenarios. The effectiveness and efficacy of the proposed method is further validated through a series of experiments. 
{
 Actually, this article only considered the case where the tensor is T-PSD, which may limit the proposed procedure of practical use. However, the analytical framework constructed here can inspire future work and, with some modifications, can be naturally extended to more general cases. Specifically, the major change moving from T-PSD tensors to general tensors is that we need to decompose the asymmetric tensor $\X\in\mathbb{R}^{n_1 \times n_2 \times n_3}$ into two factor tensors, $\L\in\mathbb{R}^{n_1 \times r \times n_3}$ and $\R\in\mathbb{R}^{n_2 \times r \times n_3}$. This could result in two challenges. First, the original analytical framework becomes inapplicable and we need to simultaneously analyze the gradient descent process for both $\L$ and $\R$, which is more complex than in the symmetric case. Specifically, we need to introduce two different subspaces based on t-SVD to represent $\L$ and $\R$, respectively. Consequently, compared to the T-PSD scenario, we need to analyze more terms in the population-sample analysis. Second, if the norms of $\L$ and $\R$ are unbalanced, it can lead to algorithmic non-convergence. This issue has also been mentioned in various studies \cite{tu2016low,ma2021beyond,xiong2023over} on low-rank matrix recovery. One possible way to address this issue is to introduce a new regularization term $\lambda\|\L^* * \L - \R^* *\R\|_F$ into the objective function and include a specific analysis of this term in our presented framework.}
 
 {Finally, we emphasize that how to address the aforementioned two challenges is under our investigation, and we are expected to present some promising results in future work.}


\begin{table}[h]
\caption{$m=10r_{\star}n_3(2n-r_{\star}),\ r_{\star}=0.3n,\ n_3=5$}
\begin{tabular}{p{0.1cm}p{0.1cm}p{0.8cm}p{0.8cm}p{0.8cm}p{0.8cm}p{0.8cm}p{0.8cm}}
\hline
   &                         & \multicolumn{2}{c}{TNN}                              & \multicolumn{2}{c}{IR-t-TNN}                            & \multicolumn{2}{c}{Ours}                               \\
$n$   & $v$                      & \multicolumn{1}{c}{error} & \multicolumn{1}{c}{time(s)} & \multicolumn{1}{c}{error} & \multicolumn{1}{c}{time(s)} & \multicolumn{1}{c}{error} & \multicolumn{1}{c}{time(s)} \\ \hline
30 & 0.3                     & 0.0435                   & 9.7994                   & 0.0435                   & 3.6334                   & \textbf{0.0367}                  & \textbf{1.6395}                   \\
   & 0.5                     & 0.0716                  & 9.7912                   & 0.0716                  &{3.6764}                   & \textbf{0.0515}                  & \textbf{1.6696}                   \\
   & \multicolumn{1}{l}{0.7} & 0.1021                    & 9.7843                   & 0.1021                    & 3.6792                   & \textbf{0.0702}                 & \textbf{1.7627}                   \\ \\
50 & 0.3                     &    0.0338                &    72.7788                      &   0.0338                        & 40.7706                         &\textbf{0.0264}                           &   \textbf{9.3127}                       \\
   & 0.5                     &   0.0563                        &  72.9363                        & 0.0563                          &   40.8795                       &     \textbf{0.0394}                       & \textbf{9.3879}                         \\
   & \multicolumn{1}{l}{0.7} &         0.0790                  &    72.6651                      & 0.0790                          &   40.6241                       &   \textbf{0.0533}                        &\textbf{9.449}                          \\  \\
70 & \multicolumn{1}{l}{0.3} &0.0285                           &  396.4822                        &0.0285                           &235.2165                          &  \textbf{0.0216}                         &\textbf{30.8286}                          \\
   & \multicolumn{1}{l}{0.5} &  0.0475                         &  396.2685                        & 0.0475                          & 234.6146                         & \textbf{0.0328}                          &  \textbf{30.854}                        \\
   & \multicolumn{1}{l}{0.7} & 0.0664                          &  394.6048                        & 0.0664                          & 233.6755                         &  \textbf{0.0443}                         & \textbf{30.1598}                         \\ \hline
\end{tabular}
\label{table:2}
\end{table}

\begin{table}[h]
\caption{$m=10r_{\star}n_3(2n-r_{\star}),\ r_{\star}=0.2n,\ n_3=5$}
\begin{tabular}{p{0.1cm}p{0.1cm}p{0.8cm}p{0.8cm}p{0.8cm}p{0.8cm}p{0.8cm}p{0.8cm}}

\hline
   &                         & \multicolumn{2}{c}{TNN}                              & \multicolumn{2}{c}{IR-t-TNN}                            & \multicolumn{2}{c}{Ours}                               \\
$n$   & $v$                      & \multicolumn{1}{c}{error} & \multicolumn{1}{c}{time(s)} & \multicolumn{1}{c}{error} & \multicolumn{1}{c}{time(s)} & \multicolumn{1}{c}{error} & \multicolumn{1}{c}{time(s)} \\ \hline
30 & 0.3                     & 0.0589                    & 7.4303                   & 0.0589                   & 3.2016                  &\textbf{0.0429}                  & \textbf{1.0736}                   \\
   & 0.5                     & 0.0984                 & 7.3765                   & 0.0984                 & 3.1742                   & \textbf{0.0515}                  &\textbf{1.1114}                    \\
   & \multicolumn{1}{l}{0.7} & 0.1379                    & 7.6562                   & 0.1379                   & 3.2838                   & \textbf{0.0788}                  & \textbf{1.1631}                  \\
   \\
50 & 0.3                     &0.0456                  &  56.4187                        &0.0456                          & 33.5054                         & \textbf{0.0309}                          & \textbf{6.0095}                         \\
   & 0.5                     & 0.0760                          &  55.8555                        & 0.0760                          & 33.1774                         & \textbf{0.0443}                          &\textbf{6.0462}                          \\
   & \multicolumn{1}{l}{0.7} &  0.1052                         & 55.4032                         &0.1051                          & 32.9169                         &\textbf{0.0581}                            &\textbf{6.0301}                          \\  \\
70 & \multicolumn{1}{l}{0.3} &0.0382                           & 308.8834                         & 0.0382                          &184.8635                          &\textbf{0.0249}                           & \textbf{18.1586}                         \\ 
   & \multicolumn{1}{l}{0.5} & 0.0641                          & 307.2197                         & 0.0641                          & 183.9251                         & \textbf{0.0368}                          & \textbf{18.7787}                         \\
   & \multicolumn{1}{l}{0.7} &  0.0898                         & 304.8521                         &  0.0898                         & 182.1509                         &  \textbf{0.0496}                         &   \textbf{18.9125}                       \\ \hline
\end{tabular}
\label{table:3}
\end{table}
\begin{table}[h]
\caption{$m=10r_{\star}n_3(2n-r_{\star}),\ r_{\star}=0.1n,\ n_3=5$}

\begin{tabular}{p{0.1cm}p{0.1cm}p{0.8cm}p{0.8cm}p{0.8cm}p{0.8cm}p{0.8cm}p{0.8cm}}
\hline
   &                         & \multicolumn{2}{c}{TNN}                              & \multicolumn{2}{c}{IR-t-TNN}                            & \multicolumn{2}{c}{Ours}                               \\
$n$   & $v$                     & \multicolumn{1}{c}{error} & \multicolumn{1}{c}{time(s)} & \multicolumn{1}{c}{error} & \multicolumn{1}{c}{time(s)} & \multicolumn{1}{c}{error} & \multicolumn{1}{c}{time(s)} \\ \hline
30 & 0.3                     & 0.1074                   & 4.6677                  & 0.1074                  & 3.0643                   &\textbf{0.0521}                  & \textbf{0.6427}                   \\
   & 0.5                     & 0.1750                  & 4.7164                   & 0.1750                  & 3.0935                   &\textbf{0.0688}                  & \textbf{0.6671}                   \\
   & \multicolumn{1}{l}{0.7} & 0.2533                    & 4.657                   & 0.2532                    & 3.0787                   & \textbf{0.0905}                 & \textbf{0.7101}                   \\ \\
50 & 0.3                     & 0.0838                 &36.054                          &0.0838                           &  21.2458                        &\textbf{0.0371}                           &\textbf{3.4385}                          \\
   & 0.5                     &0.1375                           &36.229                          &0.1374                           &21.4327
                          &\textbf{0.0504}                           &\textbf{3.5254}                          \\ 
   & \multicolumn{1}{l}{0.7} &   0.1943                        & 36.7619                         &0.1941                           &  21.7769                        &  \textbf{0.0678}                         & \textbf{3.8409}                         \\ \\
70 & \multicolumn{1}{l}{0.3} & 0.0693                          & 199.7492                        &0.0692 
                           & 109.3462                         &\textbf{0.0297}                           & \textbf{10.0213}                         \\
   & \multicolumn{1}{l}{0.5} &  0.1178                         & 200.8866                         &0.1177                           & 110.0804                         & \textbf{0.0425}                          &  \textbf{10.5745}                        \\
   & \multicolumn{1}{l}{0.7} &  0.1655                         &  201.7875                        & 0.1653                          & 110.0439                         &  \textbf{0.05585}                          & \textbf{11.1467}                         \\ \hline
\end{tabular}
\label{table:4}
\end{table}

\bibliographystyle{IEEEtran}
\bibliography{reference}
\begin{IEEEbiography}[{\includegraphics[width=1in,height=1.25in,clip,keepaspectratio]{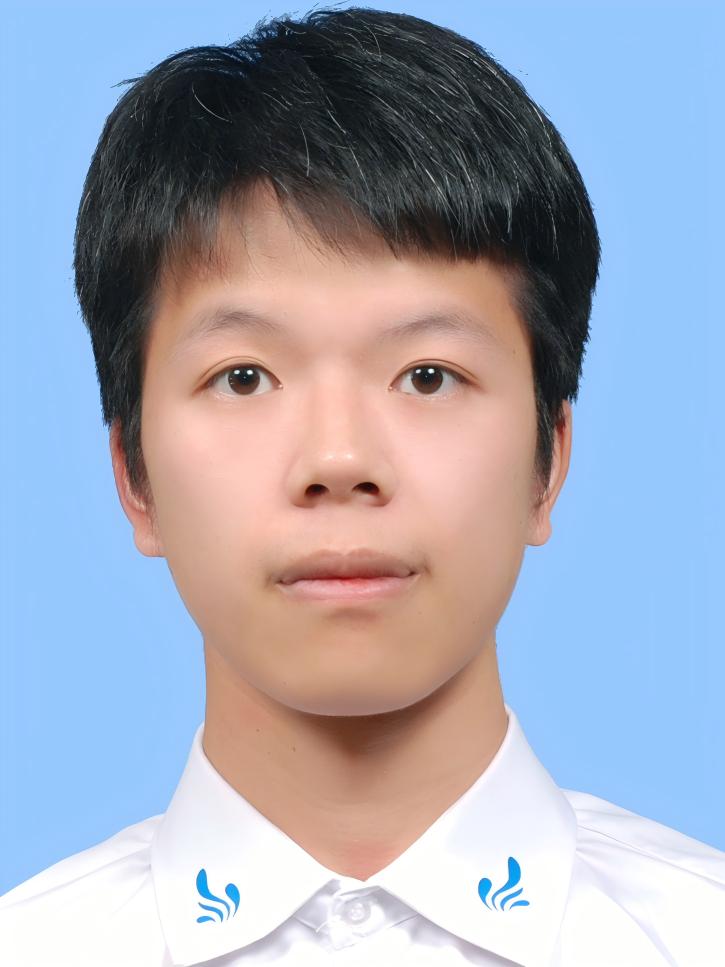}}]{Zhiyu Liu}
received the B.E. degree from Tongji University in 2021. He is currently pursuing the Ph.D. degree with the Shenyang Institute of Automation Chinese Academy of Sciences, University of Chinese Academy of Sciences. His research interests include matrix/tensor factorization and non-convex optimization.
\end{IEEEbiography}

\begin{IEEEbiography}[{\includegraphics[width=1in,height=1.25in,clip,keepaspectratio]{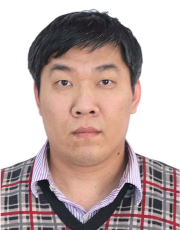}}]{Zhi Han}
received the B.S. and M.S. degrees in applied mathematics from Xi’an Jiaotong University (XJTU), Xi’an, China, in 2005 and 2007, respectively, the joint Ph.D. degree in statistics from the University of California at Los Angeles (UCLA),Los Angeles, CA, USA, in 2011, and the Ph.D.degree in applied mathematics from Xi’an Jiaotong University (XJTU) in 2012.He is currently a Professor with the State Key Laboratory of Robotics, Shenyang Institute of Automation (SIA), Chinese Academy of Sciences (CAS). His research interests include image/video modeling, low-rank matrix recovery, and deep neural networks.
\end{IEEEbiography}

\begin{IEEEbiography}[{\includegraphics[width=1in,height=1.25in,clip,keepaspectratio]{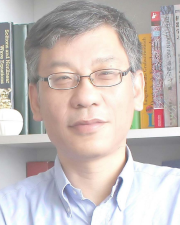}}]{Yandong Tang}
(Member, IEEE) received the B.S. and M.S. degrees in mathematics from Shandong University, China, in 1984 and 1987, respectively, and the Ph.D. degree in applied mathematics from the University of Bremen, Germany, in 2002. He is currently a Professor with the State Key Laboratory of Robotics, Shenyang Institute of Automation, Chinese Academy of Sciences. His research interests include numerical computation, image processing, and computer vision.
\end{IEEEbiography}

\begin{IEEEbiography}[{\includegraphics[width=1in,height=1.25in,clip,keepaspectratio]{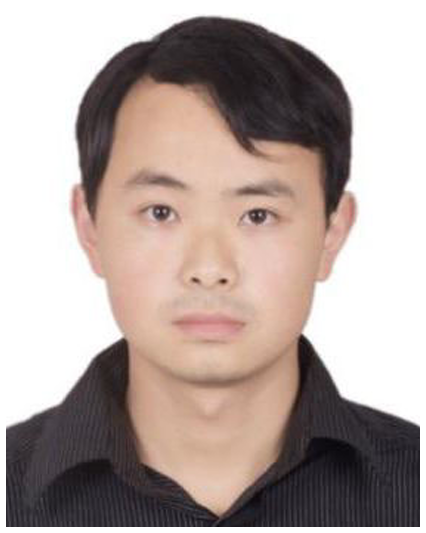}}]{Xi-Le Zhao}
received the M.S. and Ph.D. degrees from the University of Electronic Science and Tech nology of China (UESTC), Chengdu, China, in 2009 and 2012, respectively. He worked as a Post-Doctor with Prof. M. Ng at Hong Kong Baptist University, Kowloon, Hong Kong, from 2013 to 2014. He worked as a Visiting Scholar with Prof. J. Bioucas Dias at the University of Lisbon, Lisbon, Portugal, from 2016 to 2017. He is currently a Professor with the School of Mathematical Sciences, UESTC. His research interests include image processing, machine learning, and scientific computing. More information can be found on his homepage https://zhaoxile.github.io/.
\end{IEEEbiography}

\begin{IEEEbiography}[{\includegraphics[width=1in,height=1.25in,clip,keepaspectratio]{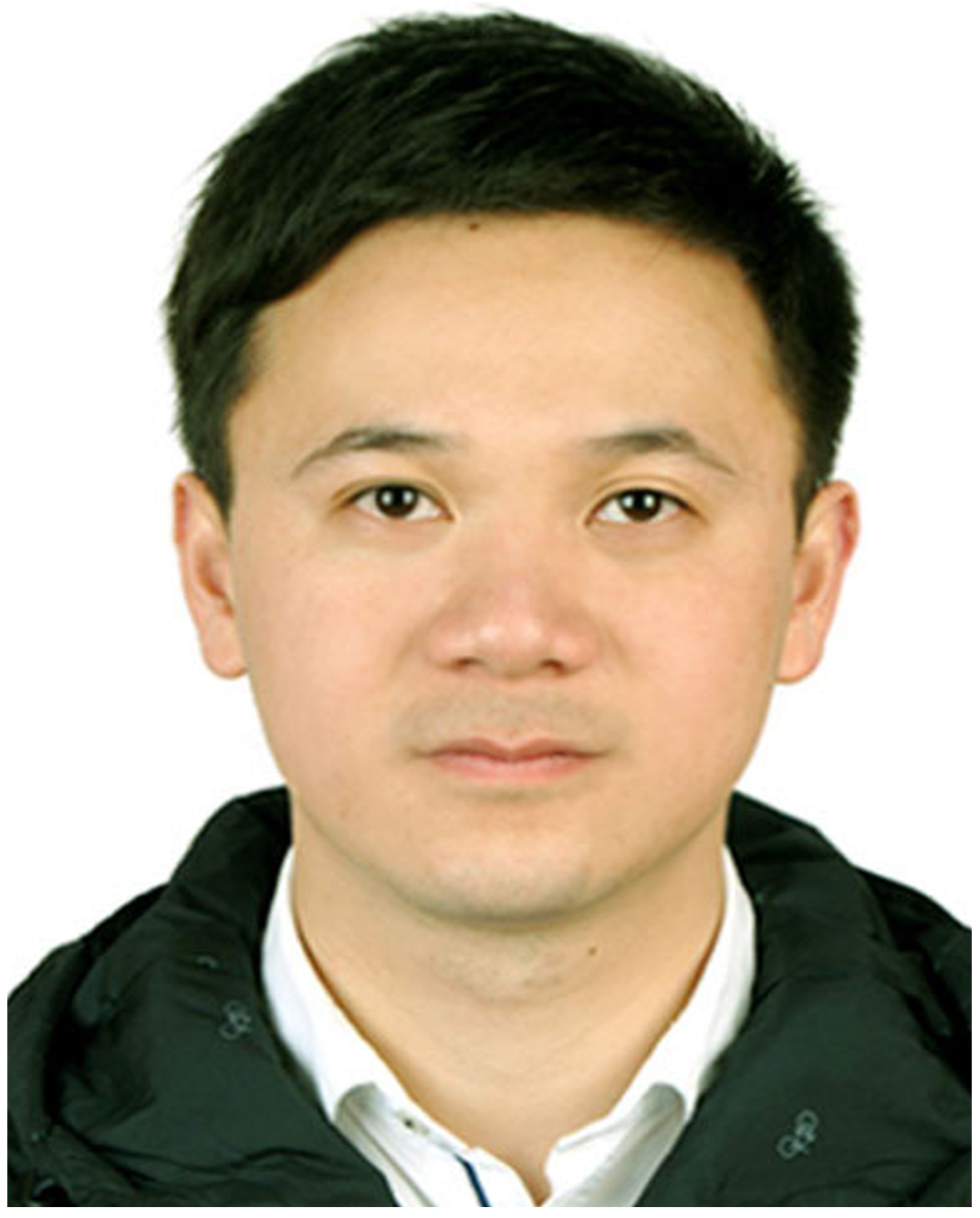}}]{Yao Wang}
received the Ph.D. degree in applied mathematics from Xi’an Jiaotong University, Xi’an, China, in 2014. He is currently an Associate Professor with the School of Management, Xi’an Jiaotong University. His current research interests include statistical signal processing, high-dimensional data analysis, and machine learning.
\end{IEEEbiography}

\end{CJK}
\end{document}